\documentclass[final]{cvpr}

\usepackage{times}
\usepackage{epsfig}
\usepackage{graphicx}
\usepackage{amsmath}
\usepackage{amssymb}
\usepackage{xcolor}
\usepackage{listings}

\usepackage[utf8]{inputenc} 
\usepackage[T1]{fontenc}    
\usepackage{url}            
\usepackage{booktabs}       
\usepackage{amsfonts}       
\usepackage{nicefrac}       
\usepackage{microtype}      

\usepackage{graphicx}
\usepackage{multirow}
\usepackage{caption}
\usepackage{amsmath}
\usepackage{amssymb}
\usepackage{xspace}
\usepackage{algorithm,algpseudocode}
\usepackage{mmstyle}
\usepackage{color}
\usepackage{subcaption}

\usepackage{stfloats}

\renewcommand{\paragraph}[1]{\noindent\textbf{#1}~~}
\usepackage[pagebackref=true,breaklinks=true,colorlinks,bookmarks=false]{hyperref}
\begin{document}
\title{Quasi-Dense Similarity Learning for Multiple Object Tracking}

\author{
    Jiangmiao Pang$^1$~~~~
    Linlu Qiu$^2$~~~~
    Xia Li$^3$~~~~
    Haofeng Chen$^4$~~~~
    Qi Li$^1$~~~~
    Trevor Darrell$^5$~~~~
    Fisher Yu$^3$~~~~
    \\
    \and
    $^1$Zhejiang University~~~~$^2$Georgia Institute of Technology
    ~~~~$^3$ETH Z{\"u}rich \\
    $^4$Stanford University~~~~$^5$UC Berkeley \\
}

\maketitle



\begin{abstract}

Similarity learning has been recognized as a crucial step for object tracking.
However, existing multiple object tracking methods only use sparse ground truth matching as the training objective, while ignoring the majority of the informative regions on the images.
In this paper, we present Quasi-Dense Similarity Learning, which densely samples hundreds of region proposals on a pair of images for contrastive learning. We can directly combine this similarity learning with existing detection methods to build Quasi-Dense Tracking (QDTrack) without turning to displacement regression or motion priors. We also find that the resulting distinctive feature space admits a simple nearest neighbor search at the inference time.
Despite its simplicity, QDTrack outperforms all existing methods on MOT, BDD100K, Waymo, and TAO tracking benchmarks.
It achieves 68.7 MOTA at 20.3 FPS on MOT17 without using external training data.
Compared to methods with similar detectors, it boosts almost 10 points of MOTA and significantly decreases the number of ID switches on BDD100K and Waymo datasets.
Our code and trained models are available at \url{http://vis.xyz/pub/qdtrack}.

\end{abstract}

\section{Introduction}
Multiple Object Tracking (MOT) is a fundamental and challenging problem in computer vision, widely used in safety monitoring, autonomous driving, video analytics, and other applications.
Contemporary MOT methods~\cite{tracktor, sort, deepsort, centertrack, jde} mainly follow the tracking-by-detection paradigm~\cite{ramanan2003finding}. That is, they detect objects on each frame and then associate them according to the estimated instance similarity.
Recent works~\cite{tracktor, sort, ioutracker, centertrack} show that if the detected objects are accurate,
the spatial proximity between objects in consecutive frames, measured by Interaction of Unions (IoUs) or center distances, is a strong prior to associate the objects.
However, this location heuristic only works well in simple scenarios.
If the objects are occluded or the scenes are crowded, this location heuristic can easily lead to mistakes.
To remedy this problem, some methods introduce motion estimation~\cite{andriyenko2011multi, choi2010multiple} or displacement regression~\cite{d&t,centertrack,chainedtracker} to ensure accurate distance estimation.

\begin{figure}
    \includegraphics[width=\linewidth]{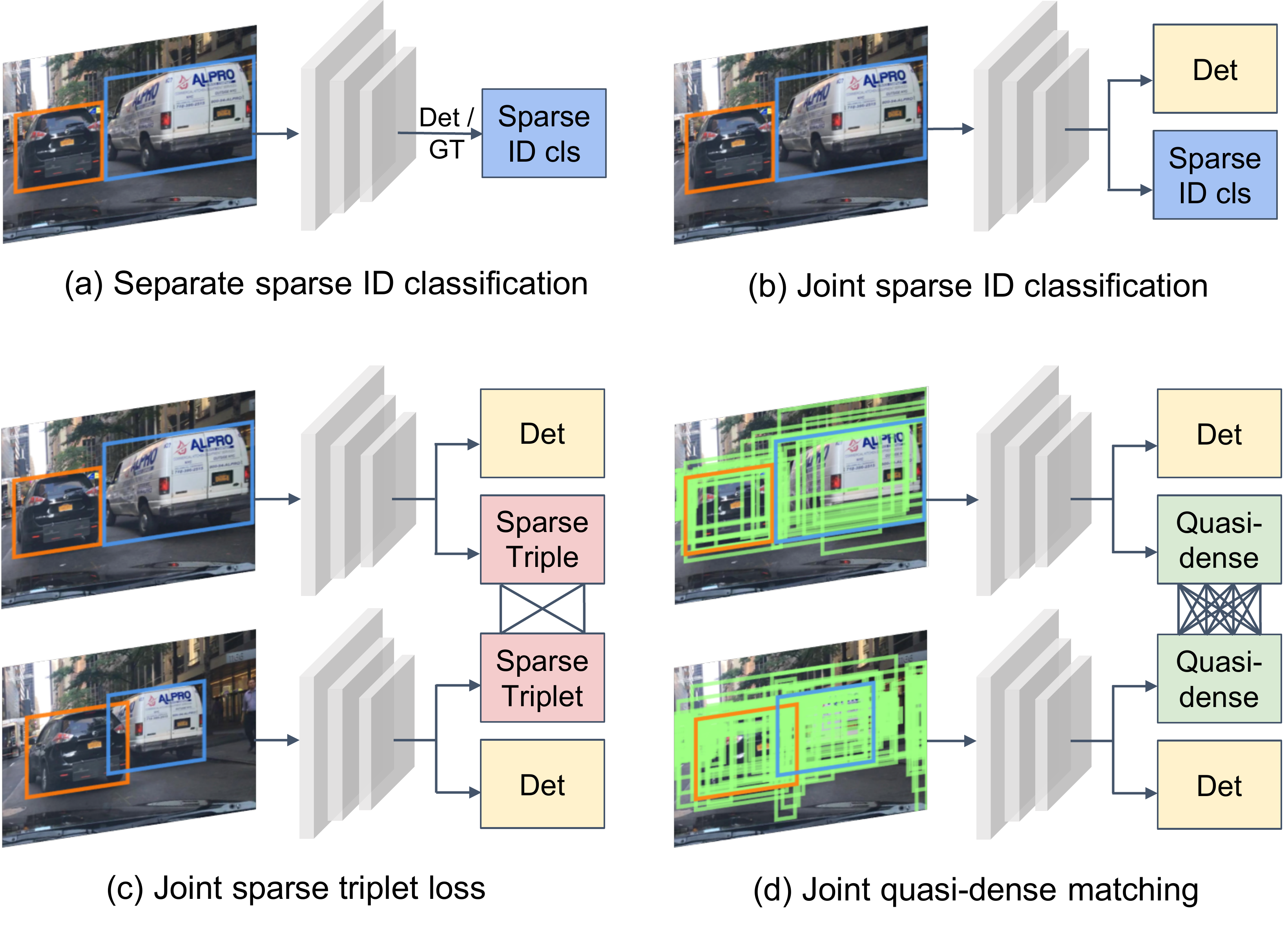}
    \caption{(a) Traditional ReID model that decouples with detector and learns with sparse ID loss; (b) joint learning ReID model with sparse ID loss; (c) joint learning ReID model with sparse triplet loss; (d) our quasi-dense similarity learning. }
    \label{fig:teaser}
\end{figure}

However, object appearance similarity usually takes a secondary role~\cite{retinatrack, deepsort} to strengthen object association or re-identify vanished objects.
The search region is constrained to be local neighborhoods to avoid distractions because the appearance features can not effectively distinguish different objects.
On the contrary, humans can easily associate the identical objects only through appearance.
We conjecture this is because the image and object information is not fully utilized for learning object similarity.
As shown in Figure~\ref{fig:teaser}, previous methods regard instance similarity learning as a post hoc stage after object detection or only use sparse ground truth bounding boxes as training samples~\cite{deepsort}.
These processes ignore the majority of the regions proposed on the images. Because objects in an image are rarely identical to each other, if the object representation is properly learned, a nearest neighbor search in the embedding space should associate and distinguish instances without bells and whistles.

We observe that besides the ground truths and detected bounding boxes, which sparsely distribute on the images, many possible object regions can provide valuable training supervision.
They are either close to the ground truth bounding boxes to provide more positive training examples or in the background as negative examples.
In this paper, we propose quasi-dense similarity learning, which densely matches hundreds of regions of interest on a pair of images for contrastive learning.
The quasi-dense samples can cover most of the informative regions on the images, providing both more box examples and hard negatives.

Because one sample has more than one positive samples on the reference image, we extend the contrastive learning~\cite{hadsell2006dimensionality, sohn2016improved, wu2018unsupervised} to multiple positive forms that makes the quasi-dense learning feasible.
Each sample is thus trained to distinguish all proposals on the other image simultaneously.
This contrast provides stronger supervision than using only the handful ground truth labels and enhances the instance similarity learning.

The inference process, which maintains the matching candidates and measures the instance similarity, also plays an important role in the tracking performance.
Besides similarity, MOT also needs to consider false positives, id switches, new appeared objects, and terminated tracks.
To tackle the missing targets with our similarity metric, we include backdrops, the unmatched objects in the last frame, for matching and use \emph{bi-directional softmax} to enforce the bi-directional consistency.
The objects that do not have matching targets will lack the consistency thus has low similarity scores to any objects.
To track the multiple targets, we also conduct duplicate removal to filter the matching candidates.

Quasi-dense similarity learning can be easily used with most existing detectors since generating region of interests is widely used in object detection algorithms.
In this paper, we apply our method to Faster R-CNN~\cite{frcnn} along with a lightweight embedding extractor and residual networks~\cite{resnet} and build \emph{Quasi-Dense Tracking} (QDTrack) models.
We conduct extensive experiments on MOT~\cite{mot16}, BDD100K~\cite{bdd100k}, Waymo~\cite{waymo}, and TAO~\cite{tao} tracking benchmarks.
Despite its simplicity, QDTrack outperforms all existing methods without bells and whistles.
It achieves 68.7 MOTA on MOT17 at 20.3 FPS without using external training data.
Moreover, it boosts almost 10 points of MOTA and significantly decreases the number of ID switches on BDD100K and Waymo datasets, establishing solid records on these brand-new large-scale benchmarks.
QDTrack allows end-to-end training, thereby simplifying the training and testing procedures of multi-object tracking frameworks.
The simplicity and effectiveness shall benefit further research.


\begin{figure*}
  \centering
  \includegraphics[width=\linewidth]{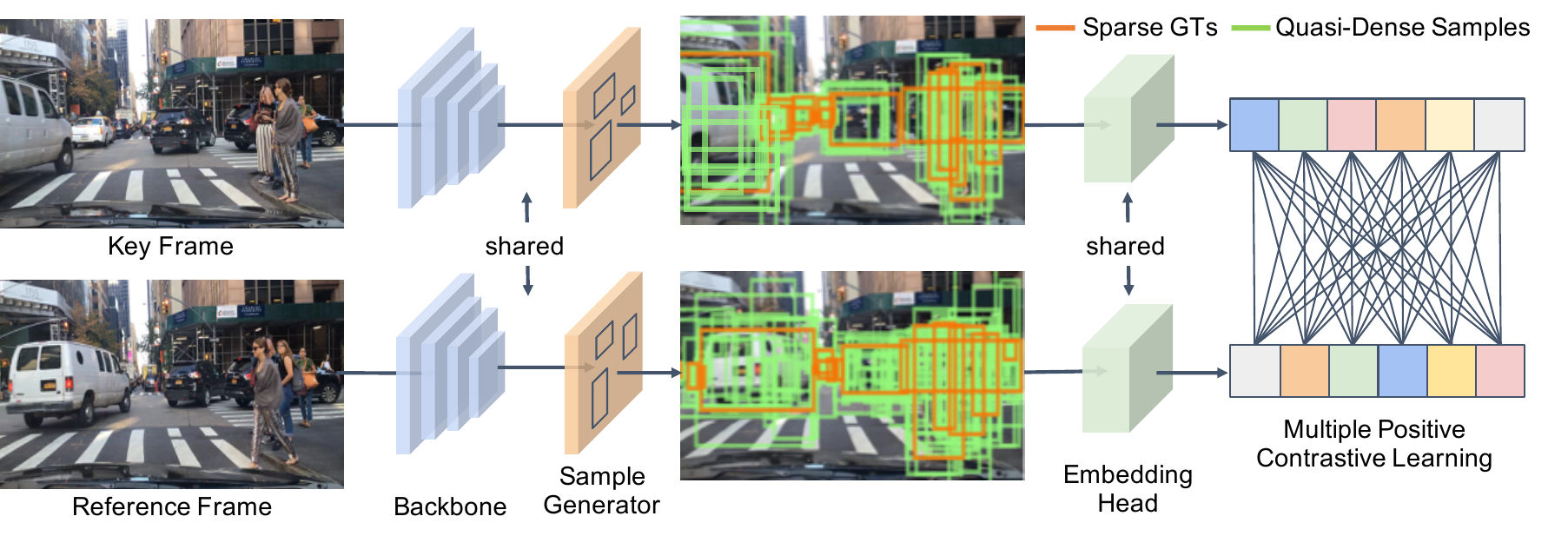}
  \caption{The training pipeline of our method. We apply dense matching between quasi-dense samples on the pair of images and optimize the network with multiple positive contrastive learning.}
  \label{fig:train}
\end{figure*}

\section{Related work}

Recent developments in multiple object tracking~\cite{leal2017tracking} follow the tracking-by-detection paradigm~\cite{ramanan2003finding}.
These approaches present different methods to estimate the instance similarity between detected objects and previous tracks, then associate objects as a bipartite matching problem~\cite{hungarian}.

\paragraph{Location and motion in MOT}
The spatial proximity has been proven effective to associate objects in consecutive frames~\cite{sort, ioutracker}.
However, they cannot do well in complicated scenarios such as crowd scenes.
Some methods use motion priors, such as Kalman Filter~\cite{sort, poi}, optical flow~\cite{xiao2018simple}, and displacement regression~\cite{goturn, d&t}, to ensure accurate distance estimations.
In contrast to the old paradigm that detects objects and predicts displacements separately, Detect \& Track \cite{d&t} is the first work that jointly optimizes object detection and tracking modules.
It predicts the displacements of the objects in consecutive frames and associates the objects with the Viterbi algorithm.
Tracktor~\cite{tracktor} directly adopts a detector as a tracker. 
CenterTrack~\cite{centertrack} and Chained-Tracker~\cite{chainedtracker} predict the object displacements with pair-wise inputs to associate the objects.
Although these methods show promising results, they~\cite{deepsort,tracktor} still need an extra re-identification model as complementary to re-identify vanished objects, making the entire framework complicated. 

\paragraph{Appearance similarity in MOT}
To exploit instance appearance similarity to strengthen tracking and re-identify vanished objects, some methods directly use an independent model~\cite{chanho2015, laura2016, crf, jeany2017, amir2017, anton2017, deepsort, tracktor} or add an extra embedding head to the detector for end-to-end training~\cite{trackrcnn, retinatrack, jde, zhang2020fair}.
However, they still learn the appearance similarity following the practice in image similarity learning, then measure the instance similarity by cosine distance.
That is, they train the model either as a $n$-classes classification problem~\cite{deepsort} where $n$ equals to the number of identities in the whole training set or using triplet loss~\cite{tripletloss}.
The classification problem is hard to extend to large-scale datasets, while the triplet loss only compares each training sample with two other identities.
These rudimentary training samples and objectives leave instance similarity learning not fully explored in MOT.
Meanwhile, they still heavily rely on motion models and displacement predictions to track objects, and the appearance similarity only takes the secondary role.

In contrast to these methods, QDTrack learns the instance similarity from dense-connected contrastive pairs and associates objects from the feature space with a simple nearest neighbor search. 
QDTrack has higher performance but with a simpler framework.
The promising results prove the power of quasi-dense similarity learning in multiple object tracking.

\paragraph{Contrastive learning}
Contrastive learning and its variants~\cite{bachman2019learning, henaff2019data, oord2018representation, tian2019contrastive, wu2018unsupervised, he2020momentum, chen2020simple, circleloss} have shown promising performance in self-supervised representation learning. 
However, it does not draw much attention when learning the instance similarity in multiple object tracking.
In this paper, we supervise dense matched quasi-dense samples with multiple positive contrastive learning by the inspiration of ~\cite{circleloss}.  
In contrast to these image-level contrastive methods, our method allows multiple positive training, while these methods can only handle the case when there is only one positive target. 
The promising results of our method shall draw the attention to contrastive learning in the multiple object tracking community.


\section{Methodology}

We propose \emph{quasi-dense similarity learning} to learn the feature embedding space that can associate identical objects and distinguish different objects for online multiple object tracking.
We define \emph{dense matching} to be matching between box candidates at all pixels, and \emph{quasi-dense} means only considering the potential object candidates at informative regions. Accordingly, \emph{sparse matching} means the method only considers ground truth labels as matching candidates when learning object association.
The main ingredients of using quasi-dense matching for multiple object tracking are object detection, instance similarity learning, and object association.

\subsection{Object detection}
Our method can be easily coupled with most existing detectors with end-to-end training.
In this paper, we take Faster R-CNN~\cite{frcnn} with Feature Pyramid Network (FPN)~\cite{fpn} as an example, while we can also apply other detectors with minor modifications.
Faster R-CNN is a two-stage detector that uses Region Proposal Network (RPN) to generate Region of Interests (RoIs). It then localizes and classifies the regions to obtain semantic labels and locations.
Based on Faster R-CNN, FPN exploits lateral connections to build the top-down feature pyramid and tackles the scale-variance problem.
The entire network is optimized with a multi-task loss function
\begin{equation}
    \mathcal{L}_\text{det} = \mathcal{L}_\text{rpn} + \lambda_1\mathcal{L}_\text{cls} + \lambda_2\mathcal{L}_\text{reg},
\end{equation}
where the RPN loss $\mathcal{L}_\text{{rpn}}$, classification loss $\mathcal{L}_\text{{cls}}$, regression loss $\mathcal{L}_\text{{reg}}$ remain the same as the original paper~\cite{frcnn}.
The loss weights $\lambda_1$ and $\lambda_2$ are set to 1.0 by default.

\subsection{Quasi-dense similarity learning}
We use the region proposals generated by RPN to learn the instance similarity with quasi-dense matching.
As shown in Figure~\ref{fig:train}, given a key image $I_1$ for training, we randomly select a reference image $I_2$ from its temporal neighborhood.
The neighbor distance is constrained by an interval $k$, where $k \in [-3, 3]$ in our experiments.
We use RPN to generate RoIs from the two images and RoI Align~\cite{maskrcnn} to obtain their feature maps from different levels in FPN according to their scales~\cite{fpn}.
We add an extra lightweight embedding head, in parallel with the original bounding box head, to extract features for each RoI.
An RoI is defined as positive to an object if they have an IoU higher than $\alpha_1$, or negative if they have an IoU lower than $\alpha_2$. $\alpha_1$ and $\alpha_2$ are 0.7 and 0.3 in our experiments.
The matching of RoIs on two frames is positive if the two regions are associated with the same object and negative otherwise.

\begin{figure*}[t]
    \centering
    \includegraphics[width=\linewidth]{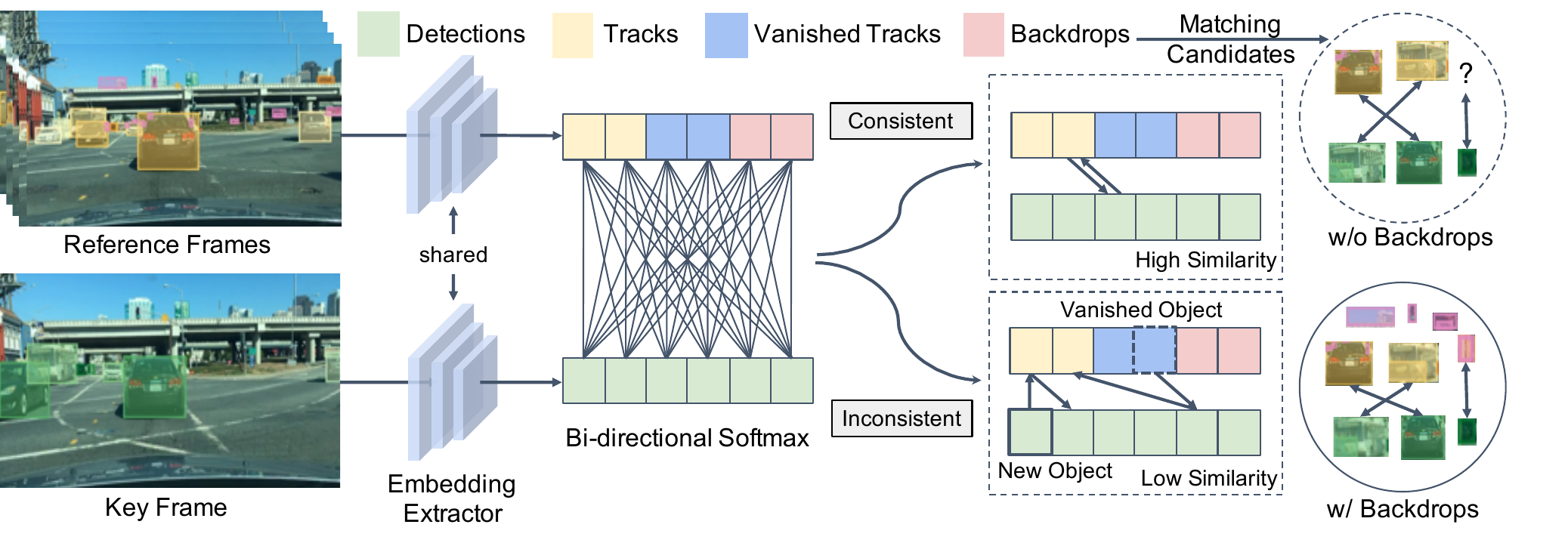}
    \caption{The testing pipeline of our method. We maintain the matching candidates and use bi-softmax to measure the instance similarity so that we can associate objects with a simple nearest neighbour search in the feature space.}
    \label{fig:test}
\end{figure*}

Assume there are $V$ samples on the key frame as training samples and $K$ samples on the reference frame as contrastive targets.
For each training sample, we can use the non-parametric softmax~\cite{wu2018unsupervised, oord2018representation} with cross-entropy to optimize the feature embeddings
\begin{align}
    \mathcal{L}_\text{embed} & = -\text{log}
    \frac{\text{exp}(\textbf{v} \cdot \textbf{k}^{+})}
    {\text{exp}(\textbf{v} \cdot \textbf{k}^{+}) + \sum_{\textbf{k}^{-}}\text{exp}(\textbf{v} \cdot \textbf{k}^{-})},
    \label{eqa:npair}
\end{align}
where $\textbf{v}$, $\textbf{k}^{+}$, $\textbf{k}^{-}$ are feature embeddings of the training sample, its positive target, and negative targets in $K$.
The overall embedding loss is averaged across all training samples, but we only illustrate one training sample for simplicity.

We apply dense matching between RoIs on the pairs of images, namely, each sample on $I_1$ is matched to all samples on $I_2$, in contrast to only using sparse sample crops, mostly ground truth boxes, to learn instance similarity in previous works~\cite{bertinetto2016fully, tripletloss}.
Each training sample on the key frame has more than one positive targets on the reference frame, so Eq.~\eqref{eqa:npair} can be extended as
\begin{align}
    \mathcal{L}_\text{embed} & = -\sum_{\textbf{k}^{+}}\text{log}
    \frac{\text{exp}(\textbf{v} \cdot \textbf{k}^{+})}
    {\text{exp}(\textbf{v} \cdot \textbf{k}^{+}) + \sum_{\textbf{k}^{-}}\text{exp}(\textbf{v} \cdot \textbf{k}^{-})}.
    \label{eqa:npair2}
\end{align}

However, this equation does not treat positive and negative targets fairly. Namely, each negative one is considered multiple times while only once for positive counterparts.
Alternatively, we can first reformulate Eq.~\eqref{eqa:npair} as
\begin{align}
    \mathcal{L}_\text{embed}
     & = \text{log}[1 + \sum_{\textbf{k}^{-}} \text{exp}(\textbf{v} \cdot \textbf{k}^{-} - \textbf{v} \cdot \textbf{k}^{+})].
\end{align}
Then in the multi-positive scenario, it can be extended by accumulating the positive term as
\begin{align}
    \mathcal{L}_\text{embed} =
    \text{log}[1 + \sum_{\textbf{k}^{+}} \sum_{\textbf{k}^{-}} \text{exp}(\textbf{v} \cdot \textbf{k}^{-} - \textbf{v} \cdot \textbf{k}^{+})].
    \label{eqa:multipos}
\end{align}
We further adopt L2 loss as an auxiliary loss
\begin{equation}
    \mathcal{L}_\text{aux} = (\frac{\textbf{v} \cdot \textbf{k}}{||\textbf{v}|| \cdot ||\textbf{k}||} - c)^2,
\end{equation}
where $c$ is 1 if the match of two samples is positive and 0 otherwise.
Note the auxiliary loss aims to constrain the logit magnitude and cosine similarity instead of improving the performance.

The entire network is joint optimized under
\begin{equation}
\label{eqa:loss}
    \mathcal{L} = \mathcal{L}_\text{det} + \gamma_1 \mathcal{L}_\text{embed} + \gamma_2 \mathcal{L}_\text{aux},
\end{equation}
where $\gamma_1$ and $\gamma_2$ are set to 0.25 and 1.0 by default in this paper.
We sample all positive pairs and three times more negative pairs to calculate the auxiliary loss.

\subsection{Object association}
Tracking objects across frames purely based on object feature embeddings is not trivial.
For example, if an object has no target or more than one target during matching, the nearest search will be ambiguous.
In other words, an object should have only one target in the matching candidates.
However, the actual tracking process is complex.
The false positives, id switches, newly appeared objects, and terminated tracks all increase the matching uncertainty.
We observe that our inference strategy, including ways of maintaining the matching candidates and measuring the instance similarity, can mitigate these problems.

\begin{table*}[t]
    \caption{Results on MOT16 and MOT17 test set with private detectors. Note that we do not use extra data for training. $\uparrow$ means higher is better, $\downarrow$ means lower is better. $^*$ means external data besides COCO and ImageNet is used.}
    \centering
    \small
    \setlength\tabcolsep{3.4mm}
    \begin{tabular}{clcccccccc}
        \toprule
        Dataset                & Method                             & MOTA $\uparrow$ & IDF1 $\uparrow$ & MOTP $\uparrow$ & MT $\uparrow$ & ML $\downarrow$ & FP $\downarrow$ & FN $\downarrow$ & IDs $\downarrow$ \\
        \midrule
        \multirow{6}{*}{MOT16} & TAP~\cite{tap}                     & 64.8            & \textbf{73.5}   & 78.7            & 292           & 164             & 12980           & 50635           & \textbf{571}     \\
                               & CNNMTT~\cite{cnnmtt}               & 65.2            & 62.2            & 78.4            & 246           & 162             & 6578            & 55896           & 946              \\
                               & POI$^*$~\cite{poi}                 & 66.1            & 65.1            & \textbf{79.5}   & 258           & 158             & \textbf{5061}   & 55914           & 3093             \\
                               & TubeTK\_POI$^*$~\cite{tubetk}      & 66.9            & 62.2            & 78.5            & 296           & \textbf{122}    & 11544           & 47502           & 1236             \\
                               & CTrackerV1~\cite{chainedtracker}   & 67.6            & 57.2            & 78.4            & 250           & 175             & 8934            & 48305           & 1897             \\
        \cmidrule{2-10}
                               & Ours                               & \textbf{69.8}   & 67.1            & 79.0            & \textbf{316}  & 150             & 9861            & \textbf{44050}  & 1097             \\
        \midrule
        \multirow{6}{*}{MOT17} & Tracktor++v2~\cite{tracktor}       & 56.3            & 55.1            & 78.8            & 498           & 831             & \textbf{8866}   & 235449          & 1987             \\
                               & Lif\_T$^*$~\cite{lift}             & 60.5            & 65.6            & 78.3            & 637           & 791             & 14966           & 206619          & \textbf{1189}    \\
                               & TubeTK$^*$~\cite{tubetk}           & 63.0            & 58.6            & 78.3            & 735           & \textbf{468}    & 27060           & 177483          & 4137             \\
                               & CTrackerV1~\cite{chainedtracker}   & 66.6            & 57.4            & 78.2            & 759           & 570             & 22284           & 160491          & 5529             \\
                               & CenterTrack$^*$~\cite{centertrack} & 67.8            & 64.7            & 78.4            & 816           & 579             & 18498           & 160332          & 3039             \\
        \cmidrule{2-10}
                               & Ours                               & \textbf{68.7}   & \textbf{66.3}   & \textbf{79.0}   & \textbf{957}  & 516             & 26589           & \textbf{146643} & 3378             \\
        \bottomrule
    \end{tabular}
    \label{tab:mot}
\end{table*}

\paragraph{Bi-directional softmax}
Our main inference strategy is bi-directional matching in the embedding space. Figure~\ref{fig:test} shows our testing pipeline.
Assume there are $N$ detected objects in frame $t$ with feature embeddings $\textbf{n}$, and $M$ matching candidates with feature embeddings $\textbf{m}$ from the past $x$ frames, the similarity $\textbf{f}$ between the objects and matching candidates is obtained by bi-directional softmax (bi-softmax):
\begin{equation}
    \textbf{f}(i, j) = [\frac{ \text{exp}(\textbf{n}_i \cdot \textbf{m}_j)}{\sum_{k=0}^{M-1} \text{exp}(\textbf{n}_i \cdot \textbf{m}_k )} +
    \frac{\text{exp}(\textbf{n}_i \cdot \textbf{m}_j)}{\sum_{k=0}^{N-1} \text{exp}(\textbf{n}_k \cdot \textbf{m}_j )}] / 2.
\end{equation}
The high score under bi-softmax will satisfy a bi-directional consistency. Namely, the two matched objects should be each other's nearest neighbor in the embedding space.
The instance similarity $\textbf{f}$ can directly associate objects with a simple nearest neighbor search.

\paragraph{No target cases}
Objects without a target in the feature space should not be matched to any candidates. Newly appeared objects, vanished tracks, and some false positives fall into this category.
The bi-softmax can tackle this problem directly, as it is hard for these objects to obtain bi-directional consistency, leading to low matching scores.
If a newly detected object has high detection confidence, it can start a new track.
Moreover, previous methods often directly drop the objects that do not match any tracks.
We argue that despite most of them are false positives, they are still useful regions that the following objects are likely to match.
We name these unmatched objects \emph{backdrops} and keep them during matching.
Experiments show that backdrops can reduce the number of false positives.

\paragraph{Multi-targets cases}
Most state-of-the-art detectors only do intra-class duplicate removal by None Maximum Suppression (NMS).
Consequently, some objects at the same locations might have different categories.
In most cases, only one of these objects is true positive while the others not.
This process can boost the object recall and contribute to a high mean Average Precision (mAP)~\cite{voc, coco}.
However, it will create duplicate feature embeddings.
To handle this issue, we do inter-class duplicate removal by NMS.
The IoU threshold for NMS is 0.7 for objects with high detection confidence (larger than 0.5) and 0.3 for objects with low detection confidence (lower than 0.5).


\section{Experiments}
We conduct experiments not only on the MOT~\cite{mot16} benchmark but also on the other brand-new large-scale benchmarks including BDD100K~\cite{bdd100k}, Waymo~\cite{waymo}, and TAO~\cite{tao}.
We hope our efforts can facilitate future multiple object tracking research to benefit from these large-scale datasets.
We also show the generalization ability of our method on BDD100K segmentation tracking benchmark.
More results, such as oracle analyses and failure case analyses are presented in the supplementary material.

\subsection{Datasets}

\paragraph{MOT Challenge}
We perform experiments on two MOT benchmarks: MOT16 and MOT17~\cite{mot16}. The dataset contains 7 videos (5,316 images) for training and 7 videos (5,919 images) for testing.
Only pedestrians are evaluated in this benchmark.
The video frame rate is 14 - 30 FPS.

\paragraph{BDD100K}
We use BDD100K~\cite{bdd100k} detection training set and tracking training set for training, and tracking validation/testing set for testing.
It annotates 8 categories for evaluation.
The detection set has 70,000 images.
The tracking set has 1,400 videos (278k images) for training, 200 videos (40k images) for validation, and 400 videos (80k images) for testing.
The images in the tracking set are annotated per 5 FPS with a 30 FPS video frame rate.

\begin{table*}[t]
    \caption{Results on BDD100K tracking validation and test set.
        Our method outperforms all methods on this benchmark.
    }
    \centering
    \resizebox{\linewidth}{!}{
        \begin{tabular}{lccccccccccc}
            \toprule
            Method                  & Split & mMOTA $\uparrow$ & mIDF1 $\uparrow$ & MOTA $\uparrow$ & IDF1 $\uparrow$ & FN $\downarrow$ & FP $\downarrow$ & ID Sw. $\downarrow$ & MT $\uparrow$  & ML $\downarrow$ & mAP $\uparrow$ \\
            \midrule
            Yu~\etal~\cite{bdd100k} & val   & 25.9             & 44.5             & 56.9            & 66.8            & 122406          & 52372           & 8315                & 8396           & 3795            & 28.1           \\
            Ours                    & val   & \textbf{36.6}    & \textbf{50.8}    & \textbf{63.5}   & \textbf{71.5}   & \textbf{108614} & \textbf{46621}  & \textbf{6262}       & \textbf{9481}  & \textbf{3034}   & \textbf{32.6}  \\
            \midrule
            Yu~\etal~\cite{bdd100k} & test  & 26.3             & 44.7             & 58.3            & 68.2            & 213220          & 100230          & 14674               & 16299          & 6017            & 27.9           \\
            DeepBlueAI              & test  & 31.6             & 38.7             & 56.9            & 56.0            & 292063          & \textbf{35401}  & 25186               & 10296          & 12266           & -              \\
            madamada                & test  & 33.6             & 43.0             & 59.8            & 55.7            & 209339          & 76612           & 42901               & 16774          & \textbf{5004}   & -              \\
            Ours                    & test  & \textbf{35.5}    & \textbf{52.3}    & \textbf{64.3}   & \textbf{72.3}   & \textbf{201041} & 80054           & \textbf{10790}      & \textbf{17353} & 5167            & \textbf{31.8}  \\
            \bottomrule
        \end{tabular}
    }
    \label{tab:bdd}
\end{table*}

\begin{table*}[t]
    \caption{Results on Waymo tracking validation set using py-motmetrics library (top) \protect \footnotemark and test set using official evaluation.
        *~indicates methods using undisclosed detectors.
    }
    \centering
    \resizebox{\linewidth}{!}{
        \begin{tabular}{lcccccccccccc}
            \toprule
            Method                                 & Split & Category & MOTA $\uparrow$    & IDF1 $\uparrow$    & FN $\downarrow$      & FP $\downarrow$      & ID Sw. $\downarrow$ & MT $\uparrow$      & ML $\downarrow$      & mAP $\uparrow$       \\
            \midrule
            IoU baseline~\cite{retinatrack}        & val   & Vehicle  & 38.25              & -                  & -                    & -                    & -                   & -                  & -                    & 45.78                \\
            Tracktor++~\cite{tracktor,retinatrack} & val   & Vehicle  & 42.62              & -                  & -                    & -                    & -                   & -                  & -                    & 42.41                \\
            RetinaTrack~\cite{retinatrack}         & val   & Vehicle  & 44.92              & -                  & -                    & -                    & -                   & -                  & -                    & 45.70                \\
            \midrule
            Ours                                   & val   & Vehicle  & 55.6               & 66.2               & 514548               & 214998               & 24309               & 17595              & 5559                 & 49.5                 \\
            Ours                                   & val   & All      & 44.0               & 56.8               & 674064               & 264886               & 30712               & 21410              & 7510                 & 40.1                 \\
            \bottomrule
            \toprule
            Method                                 & Split & Category & MOTA/L1 $\uparrow$ & FP/L1 $\downarrow$ & MisM/L1 $\downarrow$ & Miss/L1 $\downarrow$ & MOTA/L2 $\uparrow$  & FP/L2 $\downarrow$ & MisM/L2 $\downarrow$ & Miss/L2 $\downarrow$ \\
            \midrule
            Tracktor~\cite{kim2018, waymo}         & test  & Vehicle  & 34.80              & 10.61              & 14.88                & 39.71                & 28.29               & 8.63               & 12.10                & 50.98                \\
            CascadeRCNN-SORTv2*                    & test  & All      & 50.22              & 7.79               & 2.71                 & 39.28                & 44.15               & \textbf{6.94}      & 2.44                 & 46.46                \\
            HorizonMOT*                            & test  & All      & 51.01              & 7.52               & 2.44                 & \textbf{39.03}       & \textbf{45.13}      & 7.13               & 2.25                 & \textbf{45.49}       \\
            \midrule
            Ours (ResNet-50)                       & test  & All      & 49.40              & \textbf{7.41}      & 1.46                 & 41.74                & 43.88               & 7.10               & \textbf{1.31}        & 48.21                \\
            Ours (ResNet-101 + DCN)                & test  & All      & \textbf{51.18}     & 7.64               & \textbf{1.45}        & 39.73                & 45.09               & 7.20               & \textbf{1.31}        & 46.41                \\
            \bottomrule
        \end{tabular}
    }
    \label{tab:waymo}
\end{table*}

\paragraph{Waymo}
Waymo open dataset~\cite{waymo} contains images from 5 cameras associated with 5 different directions: front, front left, front right, side left, and side right.
There are 3,990 videos (790k images) for training, 1,010 videos (200k images) for validation, and 750 videos (148k images) for testing.
It annotates 3 classes for evaluation.
The videos are annotated in 10 FPS.

\paragraph{TAO}
TAO dataset~\cite{tao} annotates 482 classes in total, which are the subset of LVIS dataset~\cite{lvis}.
It has 400 videos, 216 classes in the training set, 988 videos, 302 classes in the validation set, and 1419 videos, 369 classes in the test set.
The classes in train, validation, and test sets may not overlap.
The videos are annotated in 1 FPS.
The objects in TAO are in a long-tailed distribution that half of the objects are person and 1 / 6 of the objects are car.

\subsection{Implementation details}
We use ResNet-50~\cite{resnet} as the backbone by default in this paper. We select 128 RoIs from the key frame as training samples, and 256 RoIs from the reference frame with a positive-negative ratio of 1.0 as contrastive targets.
We use IoU-balanced sampling~\cite{pang2019libra} to sample RoIs.
We use \emph{4conv-1fc} head with group normalization~\cite{wu2018group} to extract feature embeddings.
The channel number of embedding features is set to 256 by default.
We train our models with a total batch size of 16 and an initial learning rate of 0.02 for 12 epochs.
We decrease the learning rate by 0.1 after 8 and 11 epochs.

\begin{table*}[t]
    \caption{Ablation studies on quasi-dense matching and the inference strategy on the BDD100K tracking validation set. All models are comparable on detection performance. D. R. means duplicate removal. (P) means results of the class ``pedestrian''.}
    \centering
    \resizebox{\linewidth}{!}{
        \begin{tabular}{ccccccccccc}
            \toprule
            \multicolumn{2}{c}{Quasi-Dense} & \multirow{2}{*}{Metric} & \multicolumn{2}{c}{Matching candidates} & \multirow{2}{*}{MOTA $\uparrow$} & \multirow{2}{*}{IDF1 $\uparrow$} & \multirow{2}{*}{mMOTA $\uparrow$} & \multirow{2}{*}{mIDF1 $\uparrow$} & \multirow{2}{*}{MOTA(P) $\uparrow$} & \multirow{2}{*}{IDF1(P) $\uparrow$}                                  \\
            one-positive                    & multi-positive          &                                         & ~D. R.                           & Backdrops                                                                                                                                                                                                             \\
            \midrule
            -                               & -                       & \emph{cosine}                           & -                                & -                                & 60.4                              & 63.0                              & 34.0                                & 47.9                                & 37.6          & 49.7           \\
            \checkmark                      & -                       & \emph{cosine}                           & -                                & -                                & 61.5                              & 66.8                              & 35.5                                & 50.0                                & 40.5          & 52.7           \\
            -                               & \checkmark              & \emph{cosine}                           & -                                & -                                & 62.5                              & 67.8                              & 36.2                                & 50.0                                & 44.0          & 54.3           \\
            -                               & \checkmark              & \emph{bi-softmax}                       & -                                & -                                & 62.9                              & 70.0                              & 35.4                                & 48.5                                & 45.5          & 58.8           \\
            -                               & \checkmark              & \emph{bi-softmax}                       & \checkmark                       & -                                & 63.2                              & 70.1                              & 36.4                                & 50.4                                & 45.5          & 58.3           \\
            -                               & \checkmark              & \emph{bi-softmax}                       & \checkmark                       & \checkmark                       & \textbf{63.5}                     & \textbf{71.5}                     & \textbf{36.6}                       & \textbf{50.8}                       & \textbf{46.7} & \textbf{60.2}  \\
            \midrule
                                            &                         &                                         &                                  &                                  & \textbf{+3.1}                     & \textbf{+8.5}                     & \textbf{+2.6}                       & \textbf{+2.9}                       & \textbf{+9.1} & \textbf{+10.5} \\
            \bottomrule
        \end{tabular}
    }
    \label{tab:ablation}
\end{table*}

\footnotetext{\url{https://github.com/cheind/py-motmetrics}}

Here, we first talk about the common practices if not specified mentioned afterwards.
We use the original scale of the images for training and inference.
We do not use any other data augmentation methods except random horizontal flipping.
We use a model pre-trained on ImageNet for training.
When conducting online joint object detection and tracking, we initialize a new track if its detection confidence is higher than 0.8.
The backdrops are only kept for one frame.
The objects can be associated only when they are classified as the same category.

For fair comparison with recent works,
we follow the practice~\cite{jde} on MOT17 that randomly resizes and crops the longer side of the images to 1088 and does not change the aspect ratio at the training and inference time.
Other data augmentation includes random horizontal flipping and color jittering, which is the common practice in~\cite{jde,centertrack,chainedtracker}.
We do not use extra data for training except a pre-trained model from COCO.
Note that COCO is not considered as additional training data by the official rules and widely used in most methods.

On TAO, we randomly select a scale between 640 to 800 to resize the
shorter side of images during training.
At inference time, the shorter side of the images are resized to 800.
We use a LVIS~\cite{lvis} pre-trained model, consistent with the implementation of ~\cite{tao}.
However, we observe severe over-fitting problem when training on the training videos of TAO, which hurts the detection performance. So we freeze the detection model and only fine-tune the embedding head to extract instance representations.

More details such as more hyper-parameters and momentum updating are presented in the supplementary material.

\subsection{Main results}
Our method outperforms all existing methods on aforementioned benchmarks without bells and whistles.
The performance are evaluated with the official metrics.

\paragraph{MOT}
The results with private detectors on MOT16 and MOT17 benchmarks are shown in Table~\ref{tab:mot}.
Our model achieves the best MOTA of 68.7\% and IDF1 of 66.3\% on the MOT17.
We outperform the state-of-the-art tracker CenterTrack~\cite{centertrack} by 0.9 points on MOTA and 1.6 points on IDF1 respectively.
Our method does not achieve a relatively low ID Sw. because we have a higher recall.
The number of ID Sw. will likely increase when we have more tracks.
This is also why the results with public detectors, which are shown in the supplementary material, have lower IDs, because their recall are lower (FN is higher).

\begin{table}[t]
    \caption{Ablations studies on location and motion cues on the BDD100K tracking validation set.
    }
    \centering
    \resizebox{\linewidth}{!}{
        \begin{tabular}{cccccccccccc}
            \toprule
            Appearance & IoU        & Motion     & Regression & mMOTA $\uparrow$ & mIDF1 $\uparrow$ \\
            \midrule
            -          & \checkmark & -          & -          & 26.3             & 36.0             \\
            -          & \checkmark & \checkmark & -          & 27.7             & 38.5             \\
            -          & \checkmark & -          & \checkmark & 28.6             & 39.3             \\
            \midrule
            \checkmark & -          & -          & -          & \textbf{36.6}    & \textbf{50.8}    \\
            \checkmark & \checkmark & -          & -          & 36.3             & 49.8             \\
            \checkmark & \checkmark & \checkmark & -          & 36.4             & 49.9             \\
            \checkmark & \checkmark & -          & \checkmark & 36.4             & 50.1             \\
            \bottomrule
        \end{tabular}
    }
    \label{tab:bnw}
\end{table}

\paragraph{BDD100K}
The main results on BDD100K tracking validation and testing sets are in Table~\ref{tab:bdd}.
The mMOTA and mIDF1, which represent object coverage and identity consistency respectively,
are 36.6\% and 50.8\% on the validation set, and 35.5\% and 52.3\% on the testing set.
On the two sets, our method outperforms the baseline benchmark method by 10.7 points and 9.2 points in terms of mMOTA,
and 6.3 points and 7.6 points in terms of mIDF1 respectively.
We also outperform the champion of BDD100K 2020 MOT Challenge (madamada) by a large margin but with a simpler detector.
The significant advancements demonstrate that our method enables more stable object tracking.

\paragraph{Waymo}
Table~\ref{tab:waymo} shows our main results on Waymo open dataset.
We report the results on the validation set following the setup of RetinaTrack~\cite{retinatrack}, which only conduct experiments on the vehicle class.
We also report the overall performance for future comparison.
We report the results on the test set via official rules.
Our method outperforms all baselines on both validation set and test set.
We obtain a MOTA of 44.0\% and a IDF1 of 56.8\% on the validation set.
We also obtain a MOTA/L1 of 49.40\% and a MOTA/L2 of 43.88\% on the test set.
The performance of vehicle on the validation set is 10.7, 13.0, and 17.4 points higher than RetinaTrack~\cite{retinatrack}, Tracktor++~\cite{tracktor, retinatrack}, and IoU baseline~\cite{retinatrack}, respectively.
Our model with ResNet-101 and deformable convolution (DCN) has the state-of-the-art performance on the test benchmark which is on par with the champion of Waymo 2020 2D Tracking Challenge (HorizonMOT) but only with a simple single model.

\paragraph{TAO}
We obtain 16.1 points and 12.4 points of AP50 on the validation and test set, respectively.
The results are 2.9 points and 2.2 points higher than TAO's solid baseline, which are 13.2 points and 10.2 points respectively.
Although we only boost the overall performance by 2 - 3 points, we observe that we outperform the baseline by a large margin on frequent classes, that is, 38.6 points vs. 18.5 points on person.
This improvement is buried by the average across the entire hundreds of classes.
It shows that the crucial part on TAO is still how to improve the tracking on tail classes, which should be a meaningful direction for further research.
Other details are presented in the supplementary material.

\begin{figure*}[t]
    \centering
    \includegraphics[width=\linewidth]{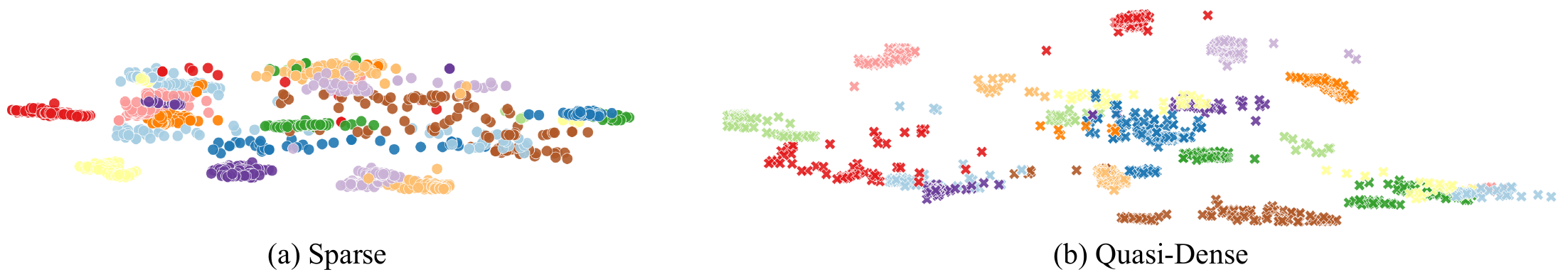}
    \caption{Visualizations of instance embeddings with (a) sparse matching and (b) quasi-dense matching using t-SNE.}
    \label{fig:embeds}
    \vspace{-4mm}
\end{figure*}

\subsection{Ablation studies}
We conduct ablation studies on BDD100K validation set, where we investigate the importance of the major model components for training and testing procedures.

\paragraph{Importance of quasi-dense matching}
The results are presented in the top sub-table of Table~\ref{tab:ablation}.
MOTA and IDF1 are calculated over all instances without considering categories as overall evaluations.
We use cosine distance to calculate the similarity scores during the inference procedure.
Compared to learning with sparse ground truths, quasi-dense tracking improves the overall IDF1 by 4.8 points (63.0\% to 67.8\%).
The significant improvement on IDF1 indicates quasi-dense tracking greatly improves the feature embeddings and enables more accurate associations.

We then analyze the improvements in detail.
In the table, we can observe that when we match each training sample to more negative samples and train the feature space with Eq.~\eqref{eqa:npair},
the IDF1 is significantly improved by 3.4 points.
This improvement contributes 70\% to the total improved 4.8 points IDF1.
This experiment shows that more contrastive targets, even most of them are negative samples, can improve the feature learning process.
The multiple-positive contrastive learning following Equation~\eqref{eqa:multipos} further improves the IDF1 by 1 point (66.8\% to 67.8\%).

\paragraph{Importance of bi-softmax}
We investigate how different inference strategies influence the performance.
As shown in the bottom part of Table~\ref{tab:ablation}, replacing cosine similarity by bi-softmax improves overall IDF1 by 2.2 points and the IDF1 of pedestrian by 4.5 points.
This experiment also shows that the one-to-one constraint further strengthens the estimated similarity.

\paragraph{Importance of matching candidates}
Duplicate removal and backdrops improve IDF1 by 1.5 points.
Overall, our training and inference strategies  improve the IDF1 by 8.5 points (63.0\% to 71.5\%).
The total number of ID switches is decreased by 30\%.
Especially, the MOTA and IDF1 of pedestrian are improved by 9.1 points and 10.5 points respectively, which further demonstrate the power of quasi-dense contrastive learning.

\paragraph{Combinations with motion and location}
Finally, we try to add the location and motion priors to understand whether they are still helpful when we have good feature embeddings for similarity measure.
These experiments follow the procedures in Tracktor~\cite{tracktor} and use the same detector for fair comparisons.
As shown in Table~\ref{tab:bnw}, without appearance features, the tracking performance is consistently improved with the introduction of additional information.
However, these cues barely enhance the performance of our approach.
Our method yields the best results when only using appearance embeddings.
The results indicate that our instance feature embeddings are sufficient for multiple object tracking with the effective quasi-dense matching, which greatly simplify the testing pipeline.

\paragraph{Inference speed}
To understand the runtime efficiency, we profile our method on NVIDIA Tesla V100.
Because it only adds a lightweight embedding head to Faster R-CNN, our method only bring marginal inference cost overhead.
With an input size of $1296 \times 720$ and a ResNet-50 backbone on BDD100K, the inference FPS is 16.4.
With an input size of $1088 \times 608$ and a ResNet-50 backbone on MOT17, the inference FPS is 20.3.

\subsection{Embedding visualizations}
We use t-SNE to visualize the embeddings trained with sparse matching and our quasi-dense matching and show them in Figure~\ref{fig:embeds}.
The instances are selected from a video in BDD100K tracking validation set.
The same instance is shown with the same color.
We observe that it is easier to separate objects in the feature space of quasi-dense matching.
More visualizations are presented in the supplementary material.

\subsection{Segmentation tracking}
We show the generalization ability of our method by extending it to instance segmentation tracking.
BDD100K provides a subset for the segmentation tracking task. There are 154 videos in the training set, 32 videos in the validation set, and 37 videos in the test set. Table \ref{tab:bddtrack} shows the results on BDD100K segmentation tracking task.
The results on the validation set are presented in the supplementary material.

\begin{table}[t]
    \caption{Results on the BDD100K segmentation tracking test set. I: ImageNet. C: COCO. S: Cityscapes. B: BDD100K. "frozen" means adopting the pretrained model from the BDD100K tracking set and only finetune the mask head.}
    \centering
    \resizebox{\linewidth}{!}{
        \begin{tabular}{lcccccc}
            \toprule
            Method           & Pretrained & mMOTSA $\uparrow$ & mMOTSP $\uparrow$ & mIDF1 $\uparrow$ & ID sw. $\downarrow$ \\
            \midrule
            SORT~\cite{sort} & I, C, S    & 12.8              & 67.3              & 28.8             & 3525                \\
            Ours             & I, C, S    & 24.0              & 66.3              & 42.5             & 1581                \\
            Ours (frozen)    & I, B       & 30.8              & 65.5              & 50.6             & 884                 \\
            \bottomrule
        \end{tabular}
    }
    \label{tab:bddtrack}
\end{table}


\section{Conclusion}
We present QDTrack, a tracking method based on quasi-dense matching for instance similarity learning.
In contrast to previous methods that use sparse ground-truth matching as similarity supervision,
we learn instance similarity from hundreds of region proposals on pairs of images, and train the feature embeddings with multiple positive contrastive learning.
In the resulting feature space, a simple nearest neighbor search can distinguish instances without bells and whistles.
Our method can be easily coupled with most of the existing detectors and trained end-to-end for multiple object tracking and segmentation tracking.

\appendix
\section{Appendix}

In this appendix, we present additional experiments, investigate oracle performance, analyze failure cases, and show some patch visualizations.
In this supplementary material, we present detailed configuration of the tracker, additional experiments and ablation studies. We also investigate oracle performance, analyze failure cases, and show some patch visualizations.

\section{Hyper-parameters}

We show the configuration of our tracker in Algorithm \ref{alg:code}.
Some of the parameters, such as number of frames to keep backdrops and the matching metric, are fixed.
For more details, please refer to our released source code.

\begin{algorithm}[h]
    \caption{Configuration of the tracker in QDTrack.}
    \label{alg:code}
    \definecolor{codeblue}{rgb}{0.25,0.5,0.5}
    \lstset{
        backgroundcolor=\color{white},
        basicstyle=\fontsize{7.2pt}{7.2pt}\ttfamily\selectfont,
        columns=fullflexible,
        breaklines=true,
        captionpos=b,
        commentstyle=\fontsize{7.2pt}{7.2pt}\color{codeblue},
        keywordstyle=\fontsize{7.2pt}{7.2pt},
    }
    \begin{lstlisting}[language=python]
tracker=dict(
    type='QuasiDenseEmbedTracker',
    # score threshold to start a new track
    init_score_thr=0.8,
    # score threshold to continue a track
    obj_score_thr=0.5,
    # score threshold for data association
    match_score_thr=0.5,
    # number of frames to keep tracks
    memo_tracklet_frames=10,
    # number of frames to keep backdrops
    memo_backdrop_frames=1,
    # momentum to update the embeddings
    memo_momentum=0.8,
    # duplicate removal to tackle multi-targets cases
    nms_backdrop_iou_thr=0.3,
    nms_class_iou_thr=0.7,
    # the matching metric
    match_metric='bisoftmax')
\end{lstlisting}
\end{algorithm}

\begin{table*}[!hb]
    \caption{Results on MOT17 test set with public detector. Note that we do not use extra data for training. $\uparrow$ means higher is better, $\downarrow$ means lower is better. $^*$ means external data besides COCO and ImageNet is used.}
    \centering
    \resizebox{\linewidth}{!}{
        \begin{tabular}{clcccccccc}
            \toprule
            Dataset                       & Method                                & MOTA $\uparrow$ & IDF1 $\uparrow$ & MOTP $\uparrow$ & MT $\uparrow$       & ML $\downarrow$     & FP $\downarrow$ & FN $\downarrow$ & IDs $\downarrow$     \\
            \midrule
            \multirow{6}{*}{Public MOT17} & Tracktor++v2~\cite{tracktor}          & 56.3            & 55.1            & 78.8            & 498 (21.1)          & 831 (35.3)          & \textbf{8866}   & 235449          & 1987 (34.1)          \\
                                          & GSM\_Tracktor~\cite{gsm}              & 56.4            & 57.8            & 77.9            & 523 (22.2)          & 813 (34.5)          & 14379           & 230174          & 1485 (25.1)          \\
                                          & MPNTrack$^*$~\cite{mpntrack}          & 58.8            & 61.7            & 78.6            & 679 (28.8)          & 788 (33.5)          & 17413           & 213594          & \textbf{1185 (19.1)} \\
                                          & Lif\_T$^*$~\cite{lift}                & 60.5            & 65.6            & 78.3            & 637 (27.0)          & 791 (33.6)          & 14966           & 206619          & 1189 (18.8)          \\
                                          & CenterTrackPub$^*$~\cite{centertrack} & 61.5            & 59.6            & 78.9            & 621 (26.4)          & 752 (31.9)          & 14076           & 200672          & 2583 (40.1)          \\
            \cmidrule{2-10}
                                          & Ours                                  & \textbf{64.6}   & \textbf{65.1}   & \textbf{79.6}   & \textbf{761 (32.3)} & \textbf{666 (28.3)} & 14103           & \textbf{182998} & 2652 (39.3)          \\
            \bottomrule
        \end{tabular}
    }
    \label{tab:mot17pub}
\end{table*}

\begin{table}[t]
    \caption{Results on TAO challenge benchmark.}
    \centering
    \resizebox{\linewidth}{!}{
        \begin{tabular}{lccccccc}
            \toprule
            Method               & Split       & AP50 & AP75 & AP  & AP50(S) & AP50(M) & AP50(L) \\
            \midrule
            SORT\_TAO~\cite{tao} & \emph{val}  & 13.2 & -    & -   & -       & -       & -       \\
            Ours                 & \emph{val}  & 16.1 & 5.0  & 7.0 & 4.8     & 13.7    & 20.0     \\
            \midrule
            SORT\_TAO~\cite{tao} & \emph{test} & 10.2 & 4.4  & 4.9 & 7.7     & 8.2     & 15.2    \\
            Ours                 & \emph{test} & 12.4 & 4.5  & 5.2 & 3.7     & 8.3     & 18.8    \\
            \bottomrule
        \end{tabular}
    }
    \label{tab:tao}
\end{table}

\begin{table}[t]
    \caption{Results on the BDD100K segmentation tracking validation set. I: ImageNet. C: COCO. S: Cityscapes. B: BDD100K. "frozen" means adopting the pretrained model from the BDD100K tracking set and only finetune the mask head.}
    \centering
    \resizebox{\linewidth}{!}{
        \begin{tabular}{lcccccc}
            \toprule
            Method           & Pretrained & mMOTSA $\uparrow$ & mMOTSP $\uparrow$ & mIDF1 $\uparrow$ & ID sw. $\downarrow$ \\
            \midrule
            SORT~\cite{sort} & I, C, S    & 11.4              & 59.7              & 22.1             & 15408               \\
            Ours             & I, C, S    & 20.2              & 59.3              & 36.0             & 1681                \\
            Ours (frozen)    & I, B       & 26.6              & 64.9              & 45.3             & 954                 \\
            \bottomrule
        \end{tabular}
    }
    \label{tab:bddtrackseg2}
\end{table}

\begin{table}[t]
    \caption{Ablation studies of momentum of the embeddings on BDD100K tracking validation set. Note the model for this table is re-trained that the results are slightly different from the results in the main paper.}
    \scriptsize
    \centering
    \resizebox{\linewidth}{!}{
        \begin{tabular}{ccccccccccc}
            \toprule
            Momentum & mMOTA $\uparrow$ & mIDF1 $\uparrow$ & MOTA $\uparrow$ & IDF1 $\uparrow$ \\
            \midrule
            0.6      & 37.0             & 50.9             & 63.3            & 71.4            \\
            0.7      & 37.0             & 50.9             & 63.3            & 71.3            \\
            0.8      & 37.0             & 50.7             & 63.3            & 71.1            \\
            0.9      & 37.0             & 50.6             & 63.3            & 70.8            \\
            1.0      & 37.0             & 50.5             & 63.3            & 70.5            \\
            \bottomrule
        \end{tabular}
    }
    \label{tab:momentum}
\end{table}

\paragraph{Dataset specific parameters.} Our object association only relies on appearance, so it is robust to different motion patterns in different datasets. The experiments share the same tracking parameters except TAO, because TAO uses 3D mAP, instead of CLEAR MOT metrics, for evaluation.

On TAO, the terms ``init\_score\_thr" and ``obj\_score\_thr" are set to 0.0001 to obtain a high recall.
Considering the numerous tracks with these thresholds, we do not maintain backdrops in these experiments.

\section{Supplementary experiments}
\paragraph{MOT17 with public detectors}
Following the strategy in Tracktor~\cite{tracktor} and CenterTrack~\cite{centertrack}, we evaluate our method with public detectors on MOT17.
That is, a new trajectory is only initialized from a public detection bounding box.
As shown in Table~\ref{tab:mot17pub}, our method outperforms existing results by a large margin.
Our method outperforms CenterTrack by 3.1 points on MOTA and 5.5 points on IDF1.

\paragraph{TAO}
Table~\ref{tab:tao} presents detailed results on the TAO~\cite{tao} dataset. Although QDTrack does not perform zero-shot and few-shot learning for the long-tail categories, our method is still a stronger baseline method on this dataset and paves the way for future studies.

\paragraph{BDD100K Segmentation Tracking}
The results on the BDD100K segmentation tracking validation set are presented in Table~\ref{tab:bddtrackseg2}.

\begin{table*}[t]
    \caption{Detection oracle analysis. The numbers in the round brackets mean the gaps between oracle results and our results.}
    \centering
    \small
    \setlength\tabcolsep{3.8mm}
    \begin{tabular}{lccccccccc}
        \toprule
        Category   & Set & MOTA $\uparrow$ & IDF1 $\uparrow$ & MOTP $\uparrow$ & FN $\downarrow$ & FP $\downarrow$ & ID Sw. $\downarrow$ & MT $\uparrow$ & ML $\downarrow$ \\
        \midrule
        Pedestrian & val & 94.3            & 79.5 (+19.3)    & 99.8            & 1               & 1               & 3226                & 3506          & 0               \\
        Rider      & val & 95.8            & 88.5 (+40.4)    & 99.9            & 0               & 0               & 107                 & 134           & 0               \\
        Car        & val & 97.7            & 86.1 (+11.1)    & 99.9            & 0               & 0               & 7716                & 13189         & 0               \\
        Bus        & val & 99.2            & 93.0 (+31.2)    & 100.0           & 0               & 0               & 72                  & 196           & 0               \\
        Truck      & val & 98.8            & 90.3 (+33.8)    & 100.0           & 0               & 0               & 340                 & 726           & 0               \\
        Bicycle    & val & 88.2            & 79.5 (+31.8)    & 98.7            & 8               & 8               & 470                 & 243           & 0               \\
        Motorcycle & val & 97.0            & 94.5 (+37.8)    & 99.8            & 0               & 0               & 27                  & 44            & 0               \\
        Train      & val & 99.4            & 98.7 (+98.7)    & 100.0           & 0               & 0               & 2                   & 6             & 0               \\
        \midrule
        All        & val & 96.3            & 88.8 (+38.0)    & 99.8            & 9               & 9               & 11960               & 18044         & 0               \\
        \bottomrule
    \end{tabular}
    \label{tab:oracle}
\end{table*}

\begin{table*}[t]
    \caption{Tracking oracle analysis. The numbers in the round brackets mean the gaps between oracle results and our results.}
    \centering
    \small
    \setlength\tabcolsep{3.6mm}
    \begin{tabular}{lccccccccc}
        \toprule
        Category   & Set & MOTA $\uparrow$ & IDF1 $\uparrow$ & MOTP $\uparrow$ & FN $\downarrow$ & FP $\downarrow$ & ID Sw. $\downarrow$ & MT $\uparrow$ & ML $\downarrow$ \\
        \midrule
        Pedestrian & val & 54.7            & 71.2 (+11.0)    & 77.6            & 14990           & 10095           & 755                 & 1835          & 367             \\
        Rider      & val & 31.4            & 52.6 (+4.5)     & 76.6            & 1390            & 242             & 115                 & 16            & 56              \\
        Car        & val & 74.3            & 82.9 (+7.9)     & 84.1            & 54585           & 31014           & 2309                & 8759          & 1141            \\
        Bus        & val & 38.2            & 65.8 (+4.0)     & 86.1            & 3532            & 2031            & 57                  & 61            & 41              \\
        Truck      & val & 37.0            & 60.9 (+4.4)     & 84.7            & 12719           & 4259            & 247                 & 149           & 239             \\
        Bicycle    & val & 30.6            & 55.6 (+7,9)     & 75.4            & 2031            & 714             & 125                 & 60            & 58              \\
        Motorcycle & val & 14.6            & 51.7 (-5.0)     & 76.4            & 443             & 292             & 35                  & 10            & 18              \\
        Train      & val & -0.6            & 0.0 (+0.0)      & 0.0             & 308             & 2               & 0                   & 0             & 6               \\
        \midrule
        All        & val & 35.0            & 55.1 (+4.3)     & 70.1            & 89998           & 48649           & 3643                & 10890         & 1926            \\
        \bottomrule
    \end{tabular}
    \label{tab:track_oracle}
\end{table*}

\section{Additional ablation studies}

\paragraph{Momentum of the embeddings.}
Assume there is an existing track and its embedding is $E_0$.
This track is associated to an object on the current frame and its embedding is $E_1$.
The new embedding of this track will be $m * E_1 + (1-m) * E_0$, where $m$ is the momentum. The momentum does not improve the results too much but it considers the history of embeddings. We show the ablation studies of different values of momentum in Table \ref{tab:momentum}.

\paragraph{Sensitivity of the $\gamma_1$ and $\gamma_2$ in Eq.~7.}
We found $\gamma_2$ does not change the final results while $\gamma_1$ does. If $\gamma_1$ is higher than 0.5, the performance will drop, but does not matter if it is lower than 0.5.

\section{Oracle analysis}

We investigate the performances of two types of oracles: detection oracle and tracking oracle on BDD100K tracking validation set.
For detection oracle, we directly extract feature embeddings of the ground truth objects in each frame and associate them using our method.
For tracking oracle, we use ground truth tracking labels to associate the detected objects.

\paragraph{Detection oracle}
The results are shown in Table~\ref{tab:oracle}.
We can observe that all MOTAs are higher than 94\%, and some of them are even close to 100\%.
This is because we use the ground truth boxes directly so that the number of false negatives and false positives are close to 0.

The metric IDF1 and ID Switches can measure the performance of identity consistency.
The average IDF1 over the 8 classes is 88.8\%, which is 38 points higher than our result.
The gaps on classes ``car'' and ``pedestrain'' are only 11.1 points and 19.3 points between oracle results and our results respectively, while gaps on other classes are exceeding 30 points.
These results show that if highly accurate detection results are provided, our method can obtain robust feature embeddings and associate objects effectively.
However, the huge performance gaps also indicate the demand of promoting detection algorithms in the video domain.
We also notice that the total number of ID switches in the oracle experiment is higher than ours.
This is due to the high object recalls in the oracle experiments, as more detected instances may introduce more ID switches accordingly.

\paragraph{Tracking oracle}
The results are shown in Table~\ref{tab:track_oracle}.
We can observe that when associating object directly with tracking labels, the mIDF1 is only boosted by 4.3 points.
This promising oracle analysis shows the effectiveness of our method and indicates that our method is bounded more by detection performance than tracking performance.

\section{Failure case analysis}
Our method can distinguish different instances even they are similar in appearance.
However, there are still some failure cases.
We show them below with figures, in which we use yellow color to represent false negatives, red color to represent false positives, and cyan color to represent ID switches.
The float number at the corner of each box indicates the detection score, while the integer indicates the object identity number.
We use green dashed box to highlight the objects we want to emphasize.

\begin{figure*}[h!]
    \centering
    \begin{subfigure}[b]{0.49\textwidth}
        \centering
        \includegraphics[width=\textwidth]{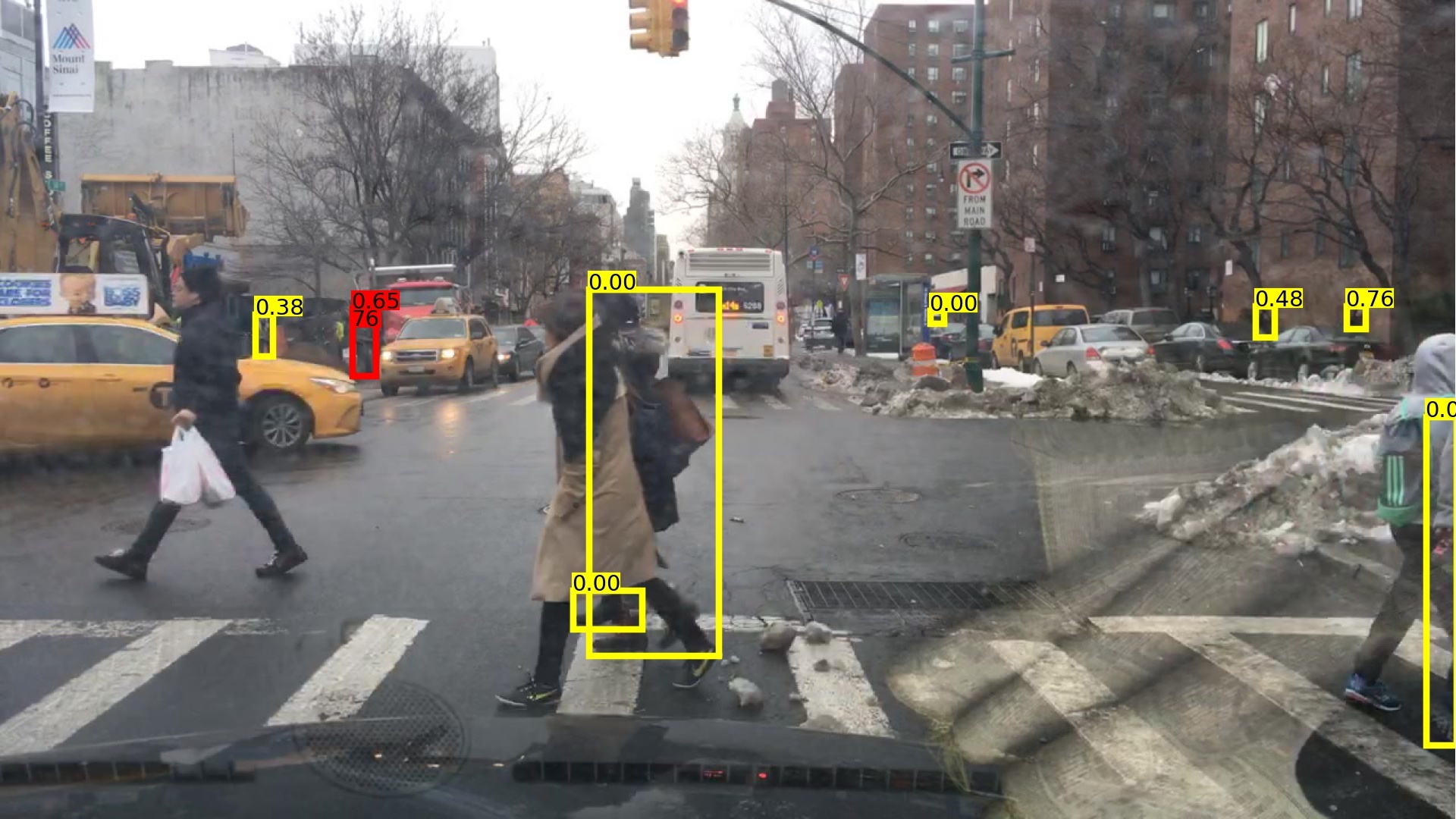}
    \end{subfigure}
    \begin{subfigure}[b]{0.49\textwidth}
        \centering
        \includegraphics[width=\textwidth]{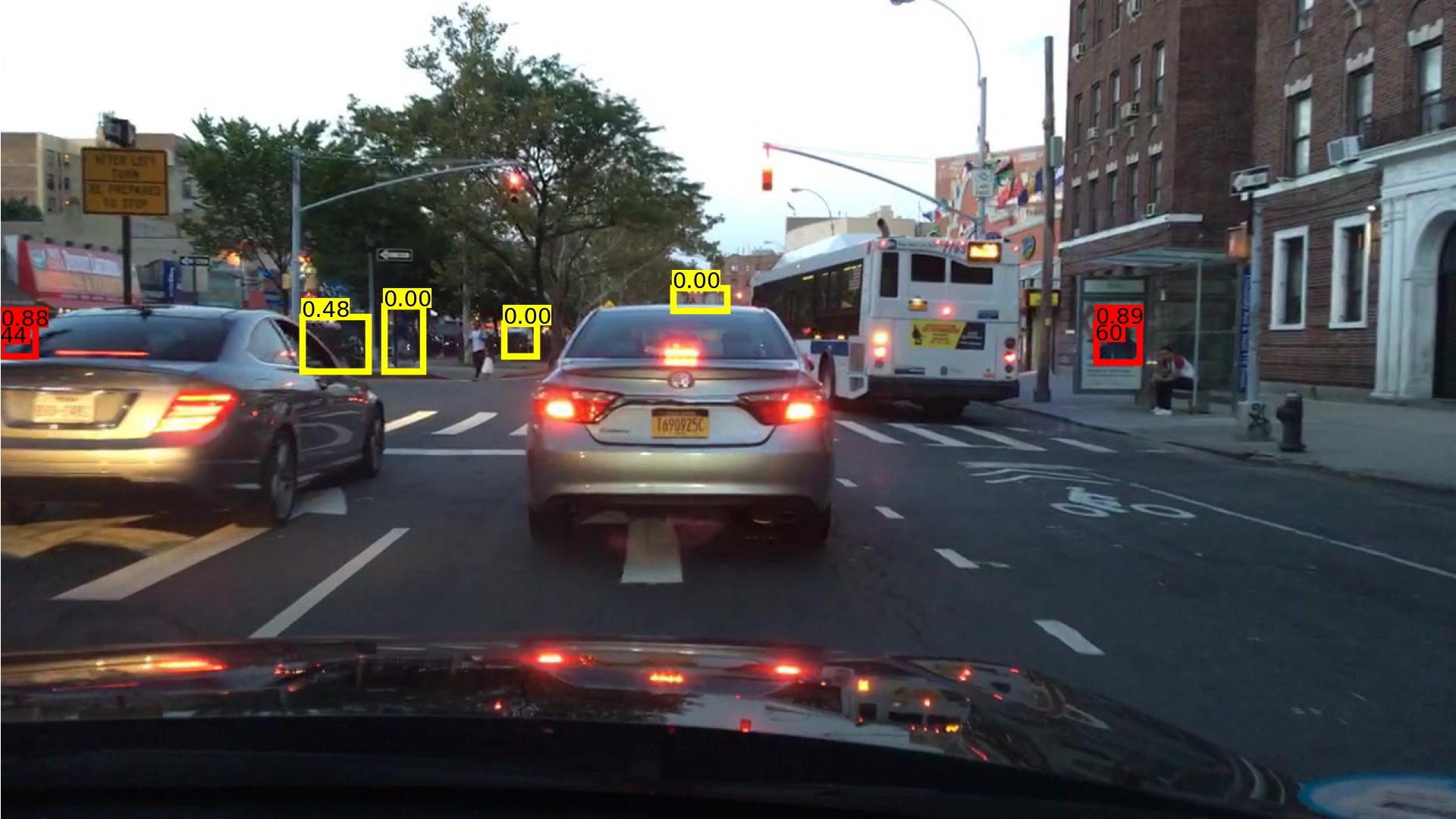}
    \end{subfigure}
    \caption{Failure cases caused by inaccurate classification confidences. The objects enclosed by yellow rectangles are false negatives, and the objects enclosed by red rectangles are false positives.}
    \label{fig:fp+fn}
\end{figure*}

\paragraph{Object classification}
Inaccurate classification confidence is the main distraction for the association procedure because false negatives and false positives destroy the one-to-one matching constraint.
As shown in Figure~\ref{fig:fp+fn}, the false negatives are mainly small objects or occluded objects under crowd scenes.
The false positives are objects that have similar appearances to annotated objects, such as persons in the mirror or advertising board, etc.

\begin{figure*}
    \centering
    \begin{subfigure}[b]{0.49\textwidth}
        \centering
        \includegraphics[width=\textwidth]{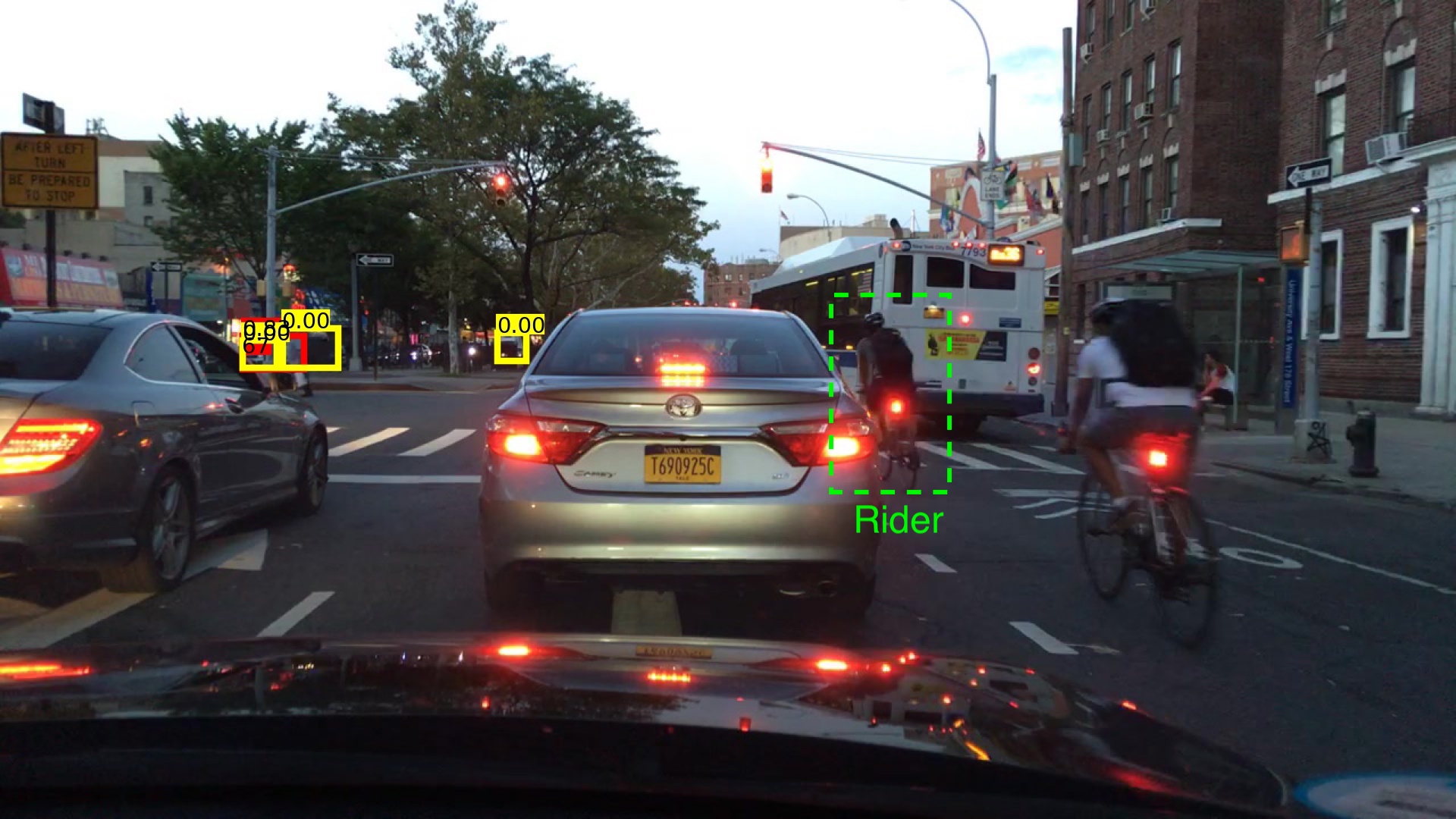}
    \end{subfigure}
    \begin{subfigure}[b]{0.49\textwidth}
        \centering
        \includegraphics[width=\textwidth]{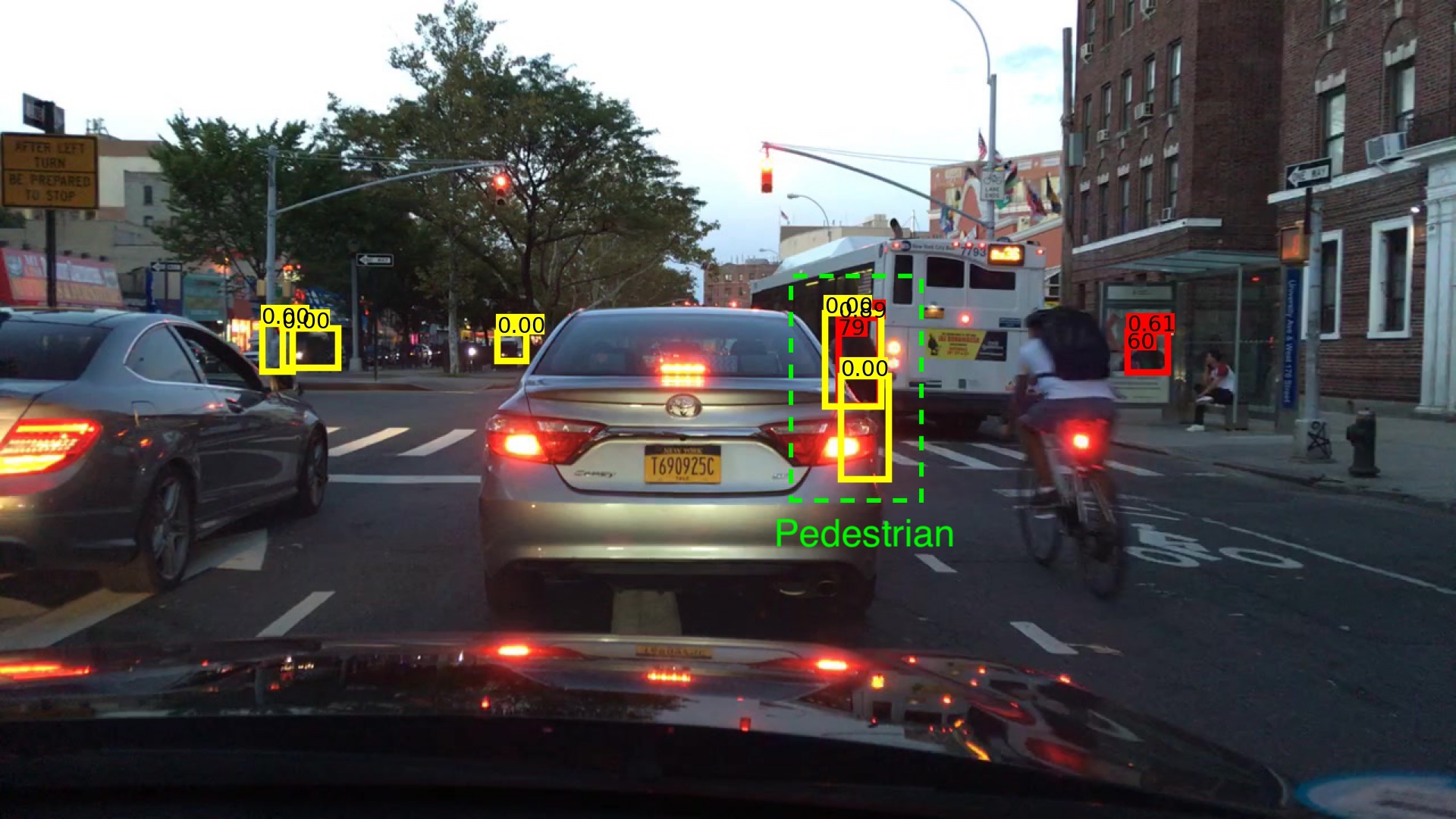}
    \end{subfigure}
    \caption{Failure case caused by inaccurate object category. The category of the highlighted object changes from ``rider'' to ``pedestrian'' due to the occlusion of the bicycle. They cannot be associated because they do not satisfy the category consistency.}
    \label{fig:det_cat}
\end{figure*}

Inaccurate object category is a less frequent distraction caused by classification.
The class of the instance may switch between different categories, which mostly belong to the same super-category.
Figure~\ref{fig:det_cat} shows an example.
The category of the highlighted object changes from ``rider'' to ``pedestrian'' when the bicycle is occluded.
Our method fails in this case because we require the associated objects have the same category.

These failure cases caused by object classification suggest the improvements on video object detection algorithms.
We can exploit temporal or tracking information to improve the detectors, thus obtaining better tracking performance.

\begin{figure*}[h!]
    \centering
    \begin{subfigure}[b]{0.49\textwidth}
        \centering
        \includegraphics[width=\textwidth]{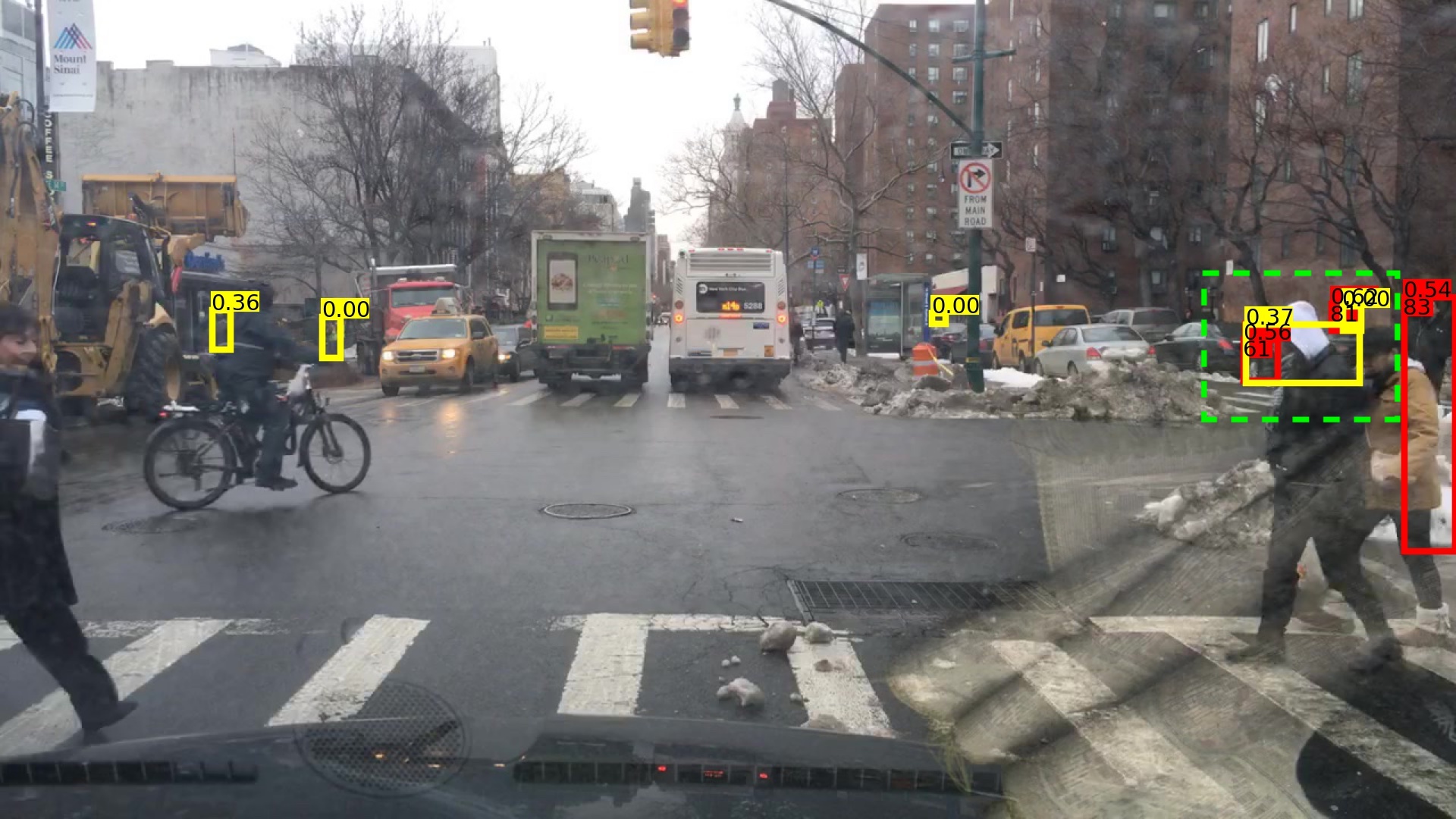}
    \end{subfigure}
    \begin{subfigure}[b]{0.49\textwidth}
        \centering
        \includegraphics[width=\textwidth]{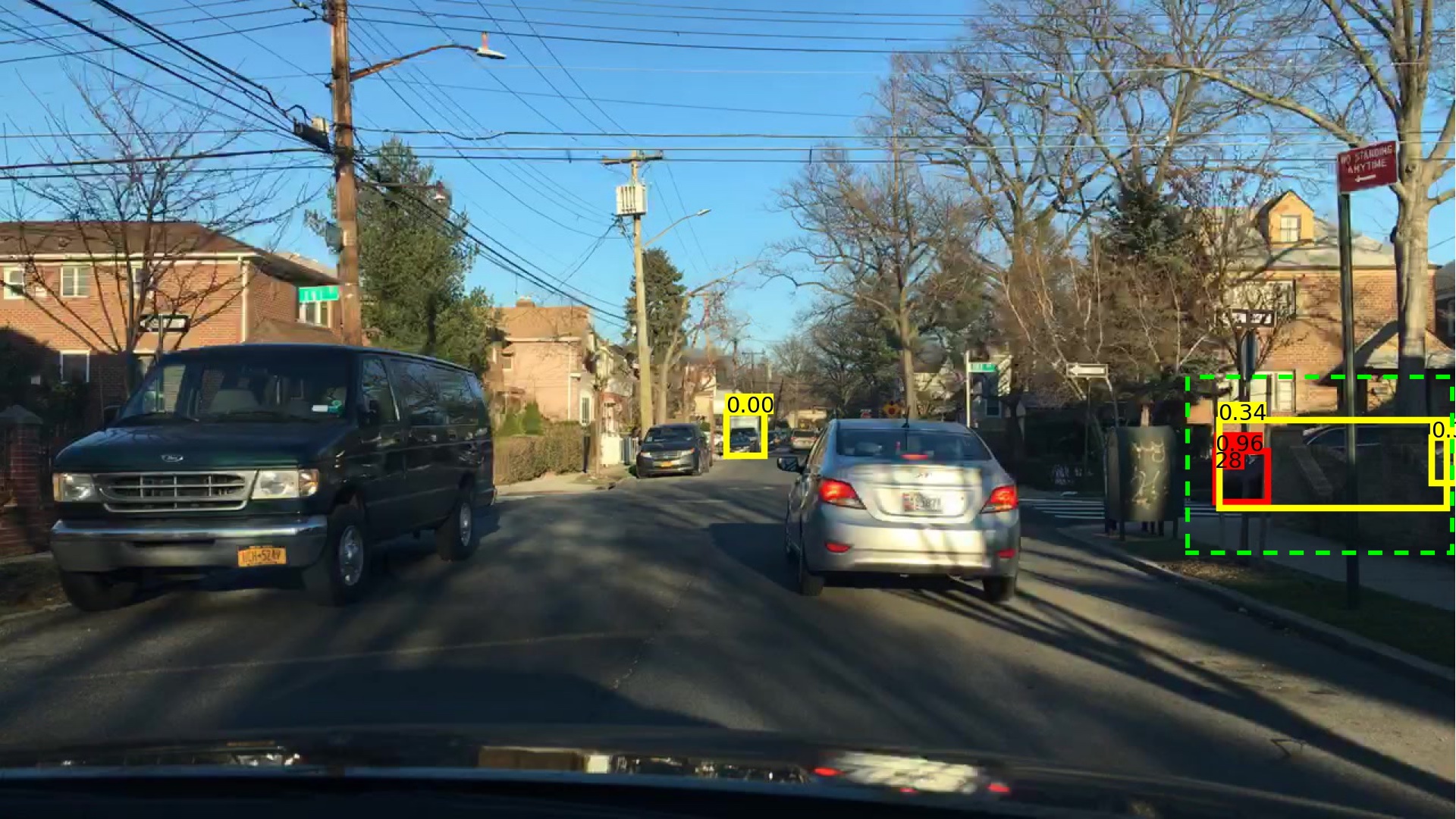}
    \end{subfigure}
    \caption{Inaccurate object localization caused by truncation. The red false positive box only covers part of the object, while the yellow box covers the entire object. They may have similar feature embeddings thus influencing the association procedure. }
    \label{fig:det_loc}
\end{figure*}

\paragraph{Object truncation/occlusion}
Object truncation/occlusion causes inaccurate object localization.
As shown in Figure~\ref{fig:det_loc}, the highlighted objects are truncated by other objects.
The detector detects two objects.
One of them is a false positive box that only covers a part of the object.
The other one is a box with a lower detection score but covers the entire object.
This case may influence the association process if the two boxes have similar feature embeddings.

\begin{figure*}[h!]
    \centering
    \begin{subfigure}[b]{0.49\textwidth}
        \centering
        \includegraphics[width=\textwidth]{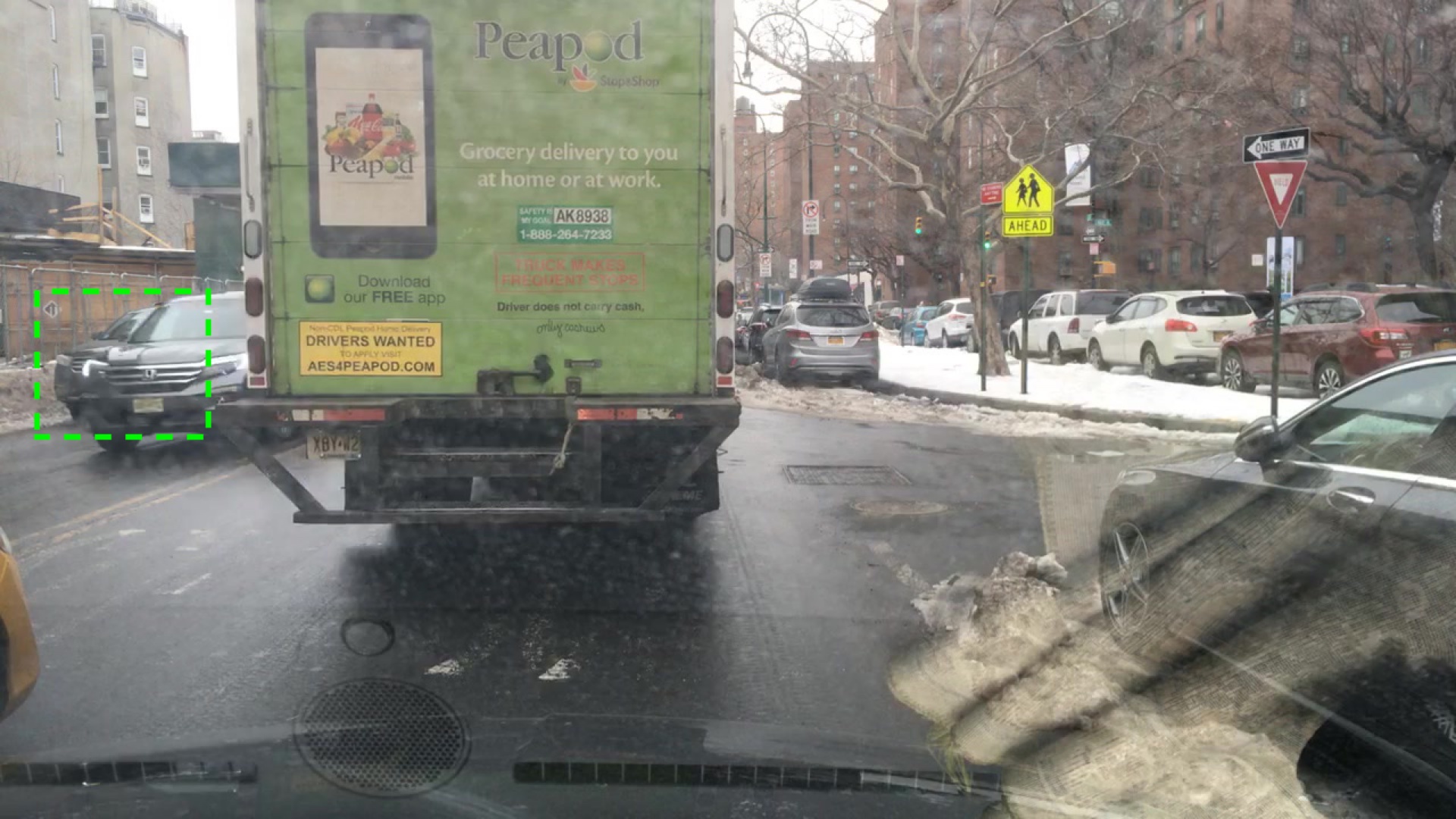}
    \end{subfigure}
    \begin{subfigure}[b]{0.49\textwidth}
        \centering
        \includegraphics[width=\textwidth]{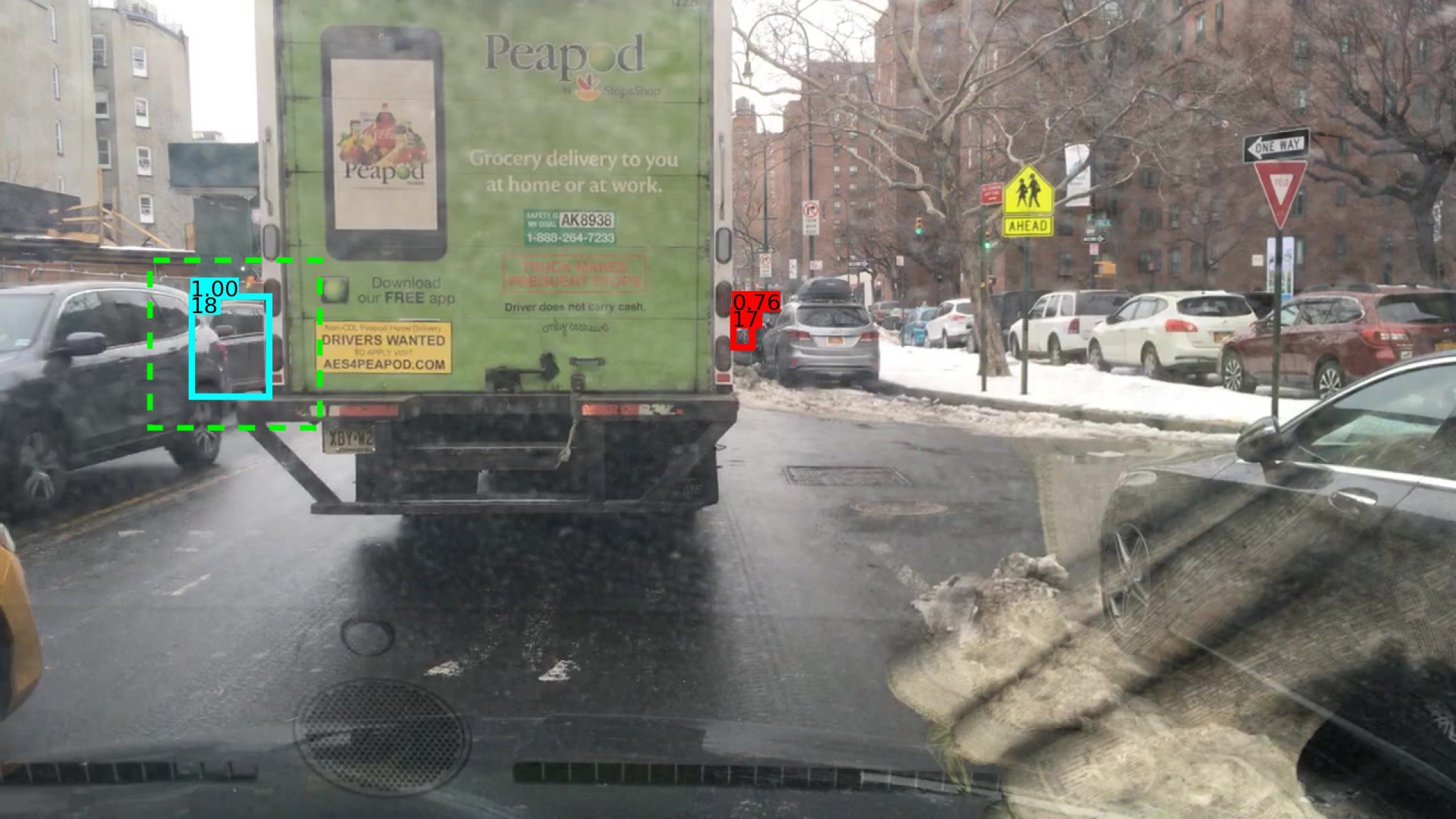}
    \end{subfigure}
    \caption{Two detected objects in different frames cover totally different regions of the object thus having low appearance similarity.}
    \label{fig:truncate2}
\end{figure*}

An instance may have totally different appearances before and after occlusion that result in low similarity scores.
As shown in Figure~\ref{fig:truncate2}, only the front of the car appears before occlusion, while only the rear of the car appears after occlusion.
Our method can associate two boxes if they cover the same discriminative regions of an object, not necessarily the exact same region.
However, if two boxes cover totally different regions of the object, they will have a low matching score.

\begin{figure*}
    \centering
    \begin{subfigure}[b]{0.49\textwidth}
        \centering
        \includegraphics[width=\textwidth]{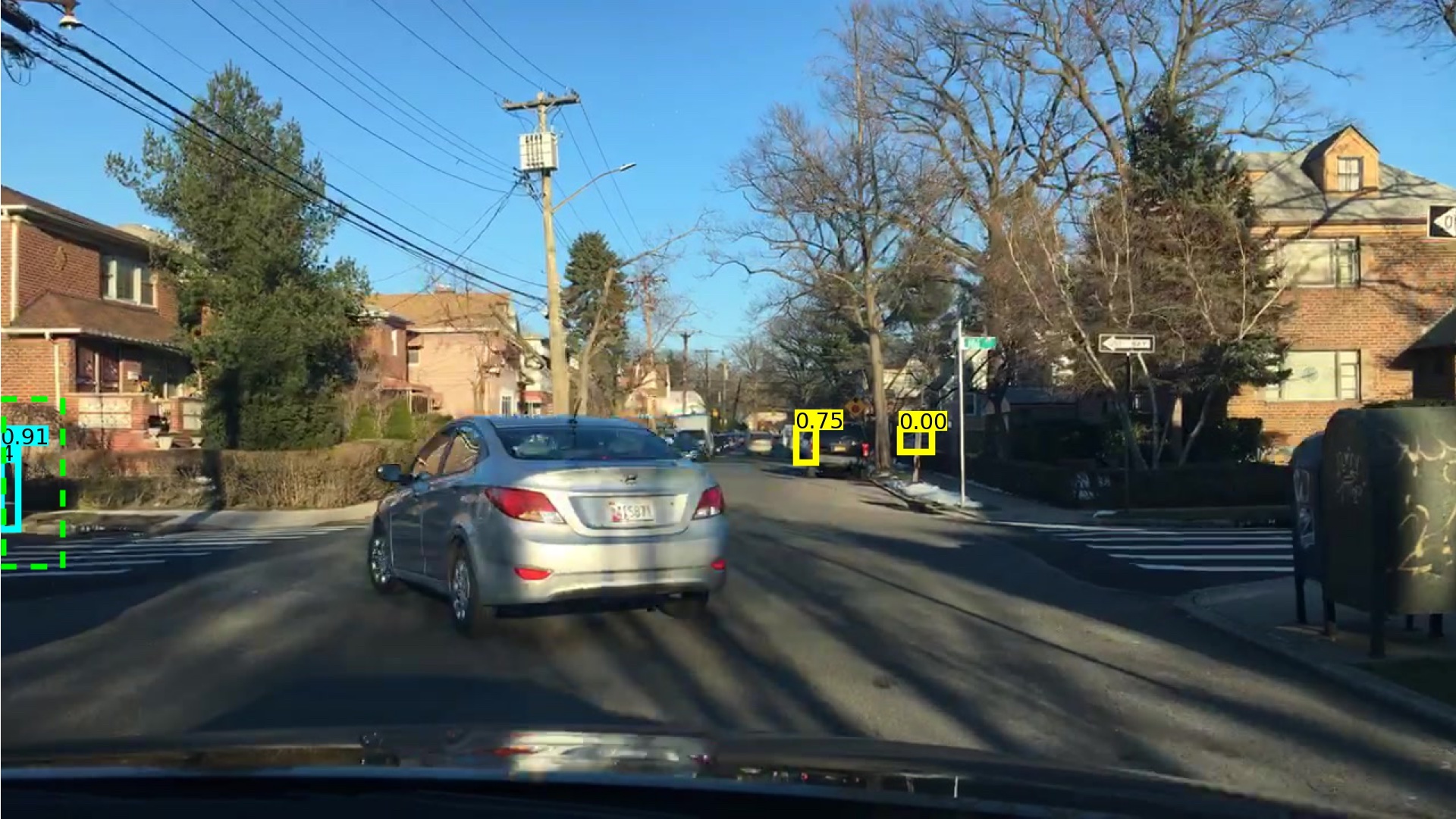}
    \end{subfigure}
    \begin{subfigure}[b]{0.49\textwidth}
        \centering
        \includegraphics[width=\textwidth]{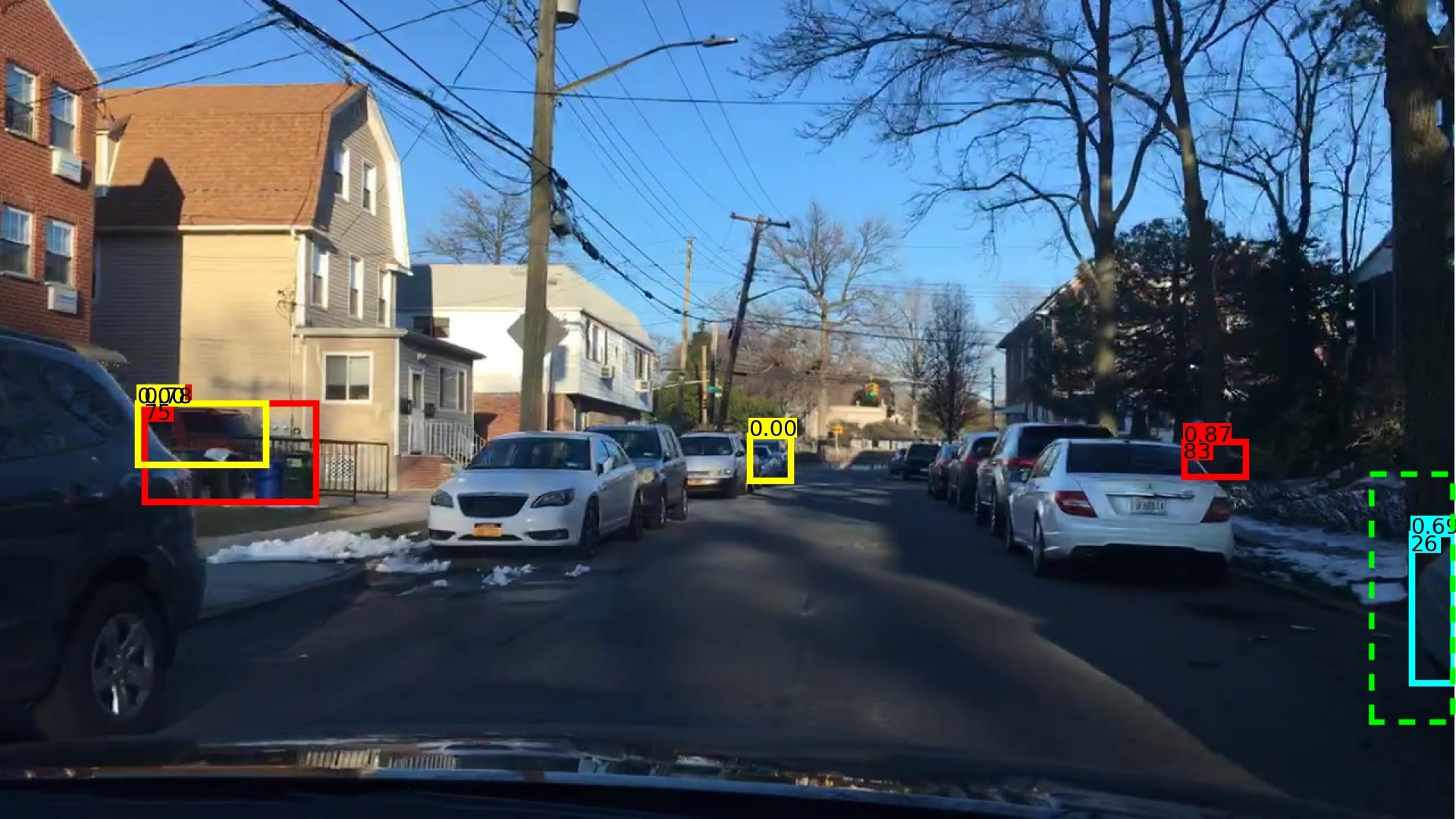}
    \end{subfigure}
    \caption{Our method cannot distinguish different instances effectively according to the limited appearance information in highly truncated objects.}
    \label{fig:truncate}
\end{figure*}

Another corner case is the extreme high-level truncation.
As shown in Figure~\ref{fig:truncate}, the highly truncated objects only appear a little when they just enter or leave the camera view.
We cannot distinguish different instances effectively according to the limited appearance information.

\begin{figure*}
    \centering
    \includegraphics[width=\linewidth]{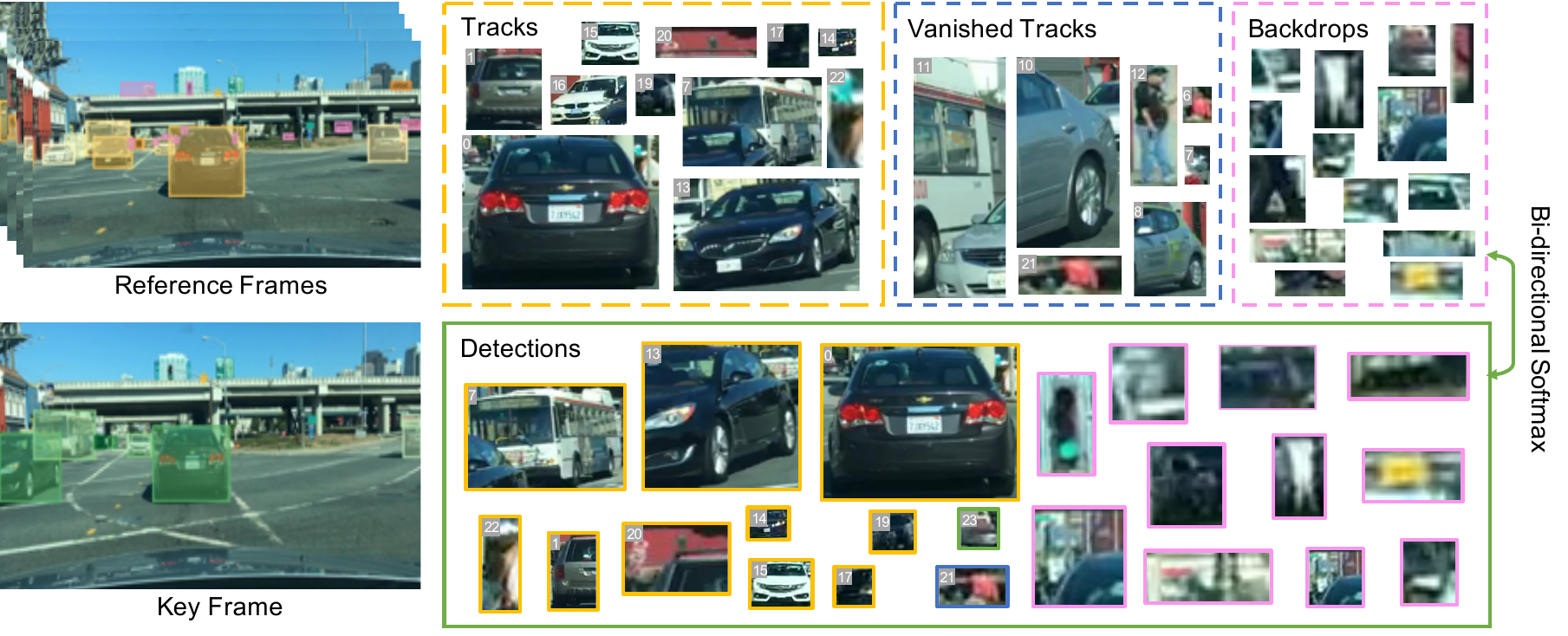}
    \caption{The visualizations of different instance patches during the testing procedure. The detected objects in the current frame are matched to tracklets in the consecutive frame, vanished tracklets, and backdrops via bi-directional softmax}
    \label{fig:test_vis}
\end{figure*}

\section{Visualizations}
We show the visualizations of different instance patches during the testing procedure in Figure~\ref{fig:test_vis}.
The detected objects in each frame are matched to prior objects via bi-directional softmax.
The prior objects include tracks in the consecutive frame, vanished tracks, and backdrops.
We annotate them with different colors.
Each detected object is enclosed by the same color of its matched object.
We can observe that most false positives in the current frame are matched to backdrops, which demonstrates keeping backdrops during the matching procedure helps reduce the number of false positives.

\section{Qualitative results}

\begin{figure*}[h!]
    \centering
    \includegraphics[width=\linewidth]{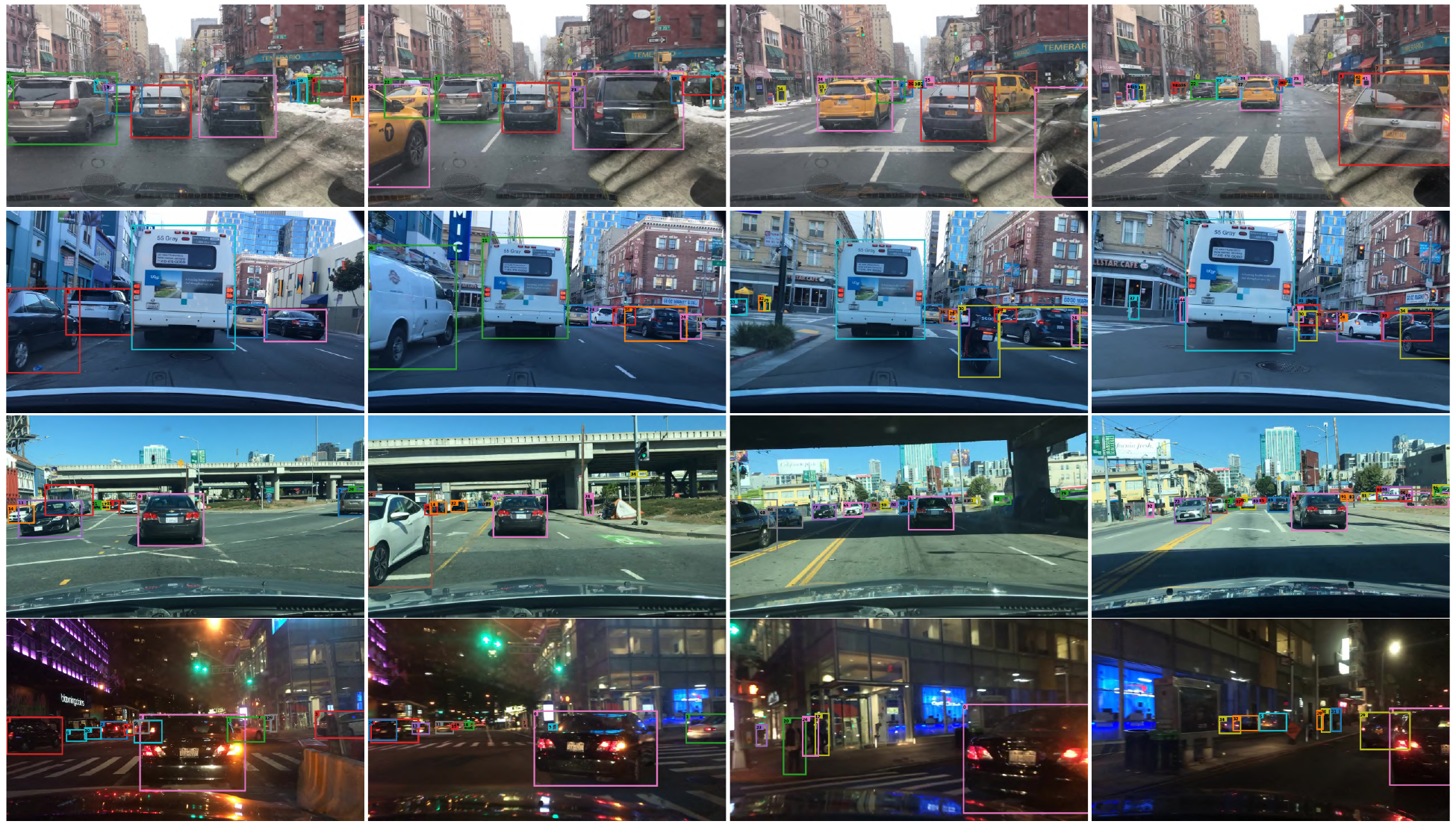}
    \caption{Qualitative results of our method on BDD100K dataset.}
    \label{fig:vis_bdd}
\end{figure*}

\begin{figure*}[h!]
    \centering
    \includegraphics[width=\linewidth]{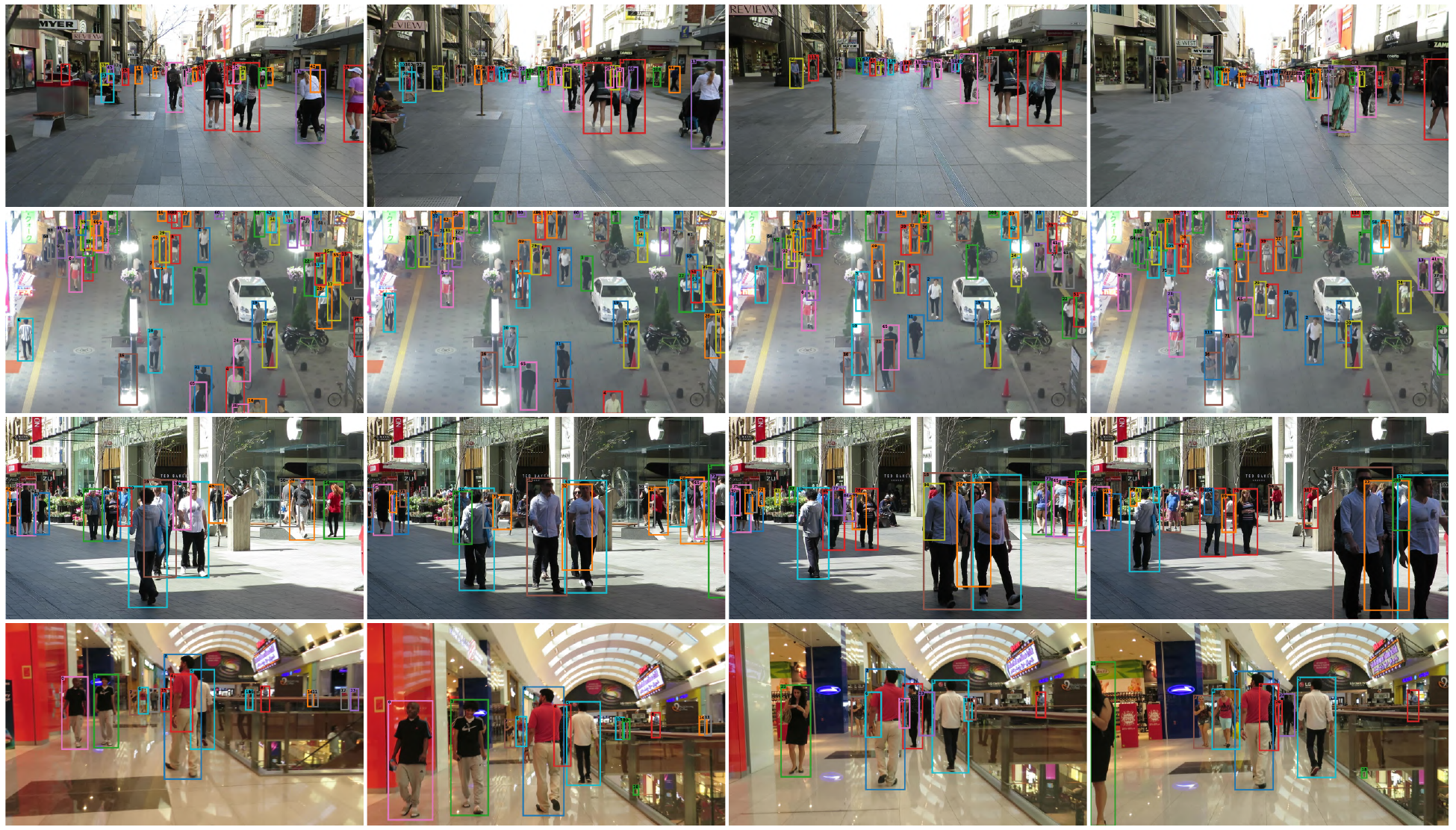}
    \caption{Qualitative results of our method on MOT17 dataset.}
    \label{fig:vis_mot}
\end{figure*}

We show some qualitative results of our method on BDD100K dataset and MOT17 dataset in Figure \ref{fig:vis_bdd} and Figure \ref{fig:vis_mot} respectively. The results are sampled from a certain interval for illustrative purposes.

{\small
    \bibliographystyle{ieee_fullname}
    \bibliography{egbib}

\begin{thebibliography}{10}\itemsep=-1pt

\bibitem{bachman2019learning}
Philip Bachman, R~Devon Hjelm, and William Buchwalter.
\newblock Learning representations by maximizing mutual information across
  views.
\newblock In {\em Advances in Neural Information Processing Systems}, 2019.

\bibitem{tracktor}
Philipp Bergmann, Tim Meinhardt, and Laura Leal{-}Taix{\'{e}}.
\newblock Tracking without bells and whistles.
\newblock {\em arXiv preprint arXiv:1903.05625}, 2019.

\bibitem{bertinetto2016fully}
Luca Bertinetto, Jack Valmadre, Joao~F Henriques, Andrea Vedaldi, and Philip~HS
  Torr.
\newblock Fully-convolutional siamese networks for object tracking.
\newblock In {\em European Conference on Computer Vision}, 2016.

\bibitem{sort}
Alex Bewley, ZongYuan Ge, Lionel Ott, Fabio~Tozeto Ramos, and Ben Upcroft.
\newblock Simple online and realtime tracking.
\newblock In {\em International Conference on Image Processing}, 2016.

\bibitem{ioutracker}
Erik Bochinski, Volker Eiselein, and Thomas Sikora.
\newblock High-speed tracking-by-detection without using image information.
\newblock In {\em IEEE International Conference on Advanced Video and Signal
  Based Surveillance (AVSS)}, 2017.

\bibitem{mpntrack}
Guillem Brasó and Laura Leal-Taixé.
\newblock Learning a neural solver for multiple object tracking.
\newblock In {\em IEEE Conference on Computer Vision and Pattern Recognition},
  2020.

\bibitem{chen2020simple}
Ting Chen, Simon Kornblith, Mohammad Norouzi, and Geoffrey Hinton.
\newblock A simple framework for contrastive learning of visual
  representations.
\newblock {\em arXiv preprint arXiv:2002.05709}, 2020.

\bibitem{choi2010multiple}
Wongun Choi and Silvio Savarese.
\newblock Multiple target tracking in world coordinate with single, minimally
  calibrated camera.
\newblock In {\em European Conference on Computer Vision}, 2010.

\bibitem{tao}
Achal Dave, Tarasha Khurana, Pavel Tokmakov, Cordelia Schmid, and Deva Ramanan.
\newblock Tao: A large-scale benchmark for tracking any object.
\newblock In {\em European Conference on Computer Vision}, 2020.

\bibitem{voc}
Mark Everingham, Luc Van~Gool, Christopher~KI Williams, John Winn, and Andrew
  Zisserman.
\newblock The pascal visual object classes (voc) challenge.
\newblock {\em International Journal of Computer Vision}, 2010.

\bibitem{d&t}
Christoph Feichtenhofer, Axel Pinz, and Andrew Zisserman.
\newblock Detect to track and track to detect.
\newblock In {\em IEEE International Conference on Computer Vision}, 2017.

\bibitem{lvis}
Agrim Gupta, Piotr Dollar, and Ross Girshick.
\newblock Lvis: A dataset for large vocabulary instance segmentation.
\newblock In {\em IEEE Conference on Computer Vision and Pattern Recognition},
  2019.

\bibitem{hadsell2006dimensionality}
Raia Hadsell, Sumit Chopra, and Yann LeCun.
\newblock Dimensionality reduction by learning an invariant mapping.
\newblock In {\em IEEE Conference on Computer Vision and Pattern Recognition},
  2006.

\bibitem{he2020momentum}
Kaiming He, Haoqi Fan, Yuxin Wu, Saining Xie, and Ross Girshick.
\newblock Momentum contrast for unsupervised visual representation learning.
\newblock In {\em IEEE Conference on Computer Vision and Pattern Recognition},
  2020.

\bibitem{maskrcnn}
Kaiming He, Georgia Gkioxari, Piotr Doll{\'a}r, and Ross Girshick.
\newblock Mask r-cnn.
\newblock In {\em IEEE International Conference on Computer Vision}, 2017.

\bibitem{resnet}
Kaiming He, Xiangyu Zhang, Shaoqing Ren, and Jian Sun.
\newblock Deep residual learning for image recognition.
\newblock In {\em IEEE Conference on Computer Vision and Pattern Recognition},
  2016.

\bibitem{goturn}
David Held, Sebastian Thrun, and Silvio Savarese.
\newblock Learning to track at 100 {FPS} with deep regression networks.
\newblock In {\em European Conference on Computer Vision}, 2016.

\bibitem{henaff2019data}
Olivier~J H{\'e}naff, Aravind Srinivas, Jeffrey De~Fauw, Ali Razavi, Carl
  Doersch, SM Eslami, and Aaron van~den Oord.
\newblock Data-efficient image recognition with contrastive predictive coding.
\newblock {\em arXiv preprint arXiv:1905.09272}, 2019.

\bibitem{tripletloss}
Alexander Hermans, Lucas Beyer, and Bastian Leibe.
\newblock In defense of the triplet loss for person re-identification.
\newblock {\em arXiv preprint arXiv:1703.07737}, 2017.

\bibitem{lift}
Andrea Hornakova, Roberto Henschel, Bodo Rosenhahn, and Paul Swoboda.
\newblock Lifted disjoint paths with application in multiple object tracking.
\newblock In {\em International Conference on Machine Learning}, 2020.

\bibitem{chanho2015}
Chanho Kim, Fuxin Li, Arridhana Ciptadi, and James~M. Rehg.
\newblock Multiple hypothesis tracking revisited.
\newblock In {\em IEEE International Conference on Computer Vision}, 2015.

\bibitem{kim2018}
Chanho Kim, Fuxin Li, and James~M. Rehg.
\newblock Multi-object tracking with neural gating using bilinear {LSTM}.
\newblock In {\em European Conference on Computer Vision}, 2018.

\bibitem{laura2016}
Laura Leal{-}Taix{\'{e}}, Cristian Canton{-}Ferrer, and Konrad Schindler.
\newblock Learning by tracking: Siamese {CNN} for robust target association.
\newblock In {\em IEEE Conference on Computer Vision and Pattern Recognition
  Workshop}, 2016.

\bibitem{leal2017tracking}
Laura Leal-Taix{\'e}, Anton Milan, Konrad Schindler, Daniel Cremers, Ian Reid,
  and Stefan Roth.
\newblock Tracking the trackers: an analysis of the state of the art in
  multiple object tracking.
\newblock {\em arXiv preprint arXiv:1704.02781}, 2017.

\bibitem{fpn}
Tsung-Yi Lin, Piotr Doll{\'a}r, Ross~B Girshick, Kaiming He, Bharath Hariharan,
  and Serge~J Belongie.
\newblock Feature pyramid networks for object detection.
\newblock In {\em IEEE Conference on Computer Vision and Pattern Recognition},
  2017.

\bibitem{coco}
Tsung-Yi Lin, Michael Maire, Serge Belongie, James Hays, Pietro Perona, Deva
  Ramanan, Piotr Doll{\'a}r, and C~Lawrence Zitnick.
\newblock Microsoft coco: Common objects in context.
\newblock In {\em European Conference on Computer Vision}, 2014.

\bibitem{gsm}
Qiankun Liu, Qi Chu, Bin Liu, and Nenghai Yu.
\newblock Gsm: Graph similarity model for multi-object tracking.
\newblock In {\em Proceedings of the Twenty-Ninth International Joint
  Conference on Artificial Intelligence}, 2020.

\bibitem{retinatrack}
Zhichao Lu, Vivek Rathod, Ronny Votel, and Jonathan Huang.
\newblock Retinatrack: Online single stage joint detection and tracking.
\newblock In {\em IEEE Conference on Computer Vision and Pattern Recognition},
  2020.

\bibitem{cnnmtt}
Nima Mahmoudi, Seyed~Mohammad Ahadi, and Mohammad Rahmati.
\newblock Multi-target tracking using cnn-based features: Cnnmtt.
\newblock {\em Multimedia Tools and Applications}, 2019.

\bibitem{mot16}
Anton Milan, Laura Leal-Taix{\'e}, Ian Reid, Stefan Roth, and Konrad Schindler.
\newblock Mot16: A benchmark for multi-object tracking.
\newblock {\em arXiv preprint arXiv:1603.00831}, 2016.

\bibitem{anton2017}
Anton Milan, Seyed~Hamid Rezatofighi, Anthony~R. Dick, Ian~D. Reid, and Konrad
  Schindler.
\newblock Online multi-target tracking using recurrent neural networks.
\newblock In {\em The AAAI Conference on Artificial Intelligence}, 2017.

\bibitem{andriyenko2011multi}
Anton Milan, Stefan Roth, and Konrad Schindler.
\newblock Continuous energy minimization for multitarget tracking.
\newblock {\em IEEE Transactions on Pattern Analysis and Machine Intelligence},
  2013.

\bibitem{hungarian}
James Munkres.
\newblock Algorithms for the assignment and transportation problems.
\newblock {\em Society for Industrial and Applied Mathematics}, 1957.

\bibitem{oord2018representation}
Aaron van~den Oord, Yazhe Li, and Oriol Vinyals.
\newblock Representation learning with contrastive predictive coding.
\newblock {\em arXiv preprint arXiv:1807.03748}, 2018.

\bibitem{tubetk}
Bo Pang, Yizhuo Li, Yifan Zhang, Muchen Li, and Cewu Lu.
\newblock Tubetk: Adopting tubes to track multi-object in a one-step training
  model.
\newblock In {\em IEEE Conference on Computer Vision and Pattern Recognition},
  2020.

\bibitem{pang2019libra}
Jiangmiao Pang, Kai Chen, Jianping Shi, Huajun Feng, Wanli Ouyang, and Dahua
  Lin.
\newblock Libra r-cnn: Towards balanced learning for object detection.
\newblock In {\em IEEE Conference on Computer Vision and Pattern Recognition},
  2019.

\bibitem{chainedtracker}
Jinlong Peng, Changan Wang, Fangbin Wan, Yang Wu, Yabiao Wang, Ying Tai,
  Chengjie Wang, Jilin Li, Feiyue Huang, and Yanwei Fu.
\newblock Chained-tracker: Chaining paired attentive regression results for
  end-to-end joint multiple-object detection and tracking.
\newblock In {\em European Conference on Computer Vision}, 2020.

\bibitem{ramanan2003finding}
Deva Ramanan and David~A Forsyth.
\newblock Finding and tracking people from the bottom up.
\newblock In {\em IEEE Conference on Computer Vision and Pattern Recognition},
  2003.

\bibitem{frcnn}
Shaoqing Ren, Kaiming He, Ross Girshick, and Jian Sun.
\newblock Faster r-cnn: Towards real-time object detection with region proposal
  networks.
\newblock In {\em Advances in Neural Information Processing Systems}, 2015.

\bibitem{amir2017}
Amir Sadeghian, Alexandre Alahi, and Silvio Savarese.
\newblock Tracking the untrackable: Learning to track multiple cues with
  long-term dependencies.
\newblock In {\em IEEE International Conference on Computer Vision}, 2017.

\bibitem{sohn2016improved}
Kihyuk Sohn.
\newblock Improved deep metric learning with multi-class n-pair loss objective.
\newblock In {\em Advances in Neural Information Processing Systems}, 2016.

\bibitem{jeany2017}
Jeany Son, Mooyeol Baek, Minsu Cho, and Bohyung Han.
\newblock Multi-object tracking with quadruplet convolutional neural networks.
\newblock In {\em IEEE Conference on Computer Vision and Pattern Recognition},
  2017.

\bibitem{waymo}
Pei Sun, Henrik Kretzschmar, Xerxes Dotiwalla, Aurelien Chouard, Vijaysai
  Patnaik, Paul Tsui, James Guo, Yin Zhou, Yuning Chai, Benjamin Caine, Vijay
  Vasudevan, Wei Han, Jiquan Ngiam, Hang Zhao, Aleksei Timofeev, Scott
  Ettinger, Maxim Krivokon, Amy Gao, Aditya Joshi, Yu Zhang, Jonathon Shlens,
  Zhifeng Chen, and Dragomir Anguelov.
\newblock Scalability in perception for autonomous driving: Waymo open dataset,
  2019.

\bibitem{circleloss}
Yifan Sun, Changmao Cheng, Yuhan Zhang, Chi Zhang, Liang Zheng, Zhongdao Wang,
  and Yichen Wei.
\newblock Circle loss: {A} unified perspective of pair similarity optimization.
\newblock {\em arXiv preprint arXiv:2002.10857}, 2020.

\bibitem{tian2019contrastive}
Yonglong Tian, Dilip Krishnan, and Phillip Isola.
\newblock Contrastive multiview coding.
\newblock {\em arXiv preprint arXiv:1906.05849}, 2019.

\bibitem{jde}
Zhongdao Wang, Liang Zheng, Yixuan Liu, and Shengjin Wang.
\newblock Towards real-time multi-object tracking.
\newblock {\em arXiv preprint arXiv:1909.12605}, 2019.

\bibitem{deepsort}
Nicolai Wojke, Alex Bewley, and Dietrich Paulus.
\newblock Simple online and realtime tracking with a deep association metric.
\newblock In {\em International Conference on Image Processing}, 2017.

\bibitem{wu2018group}
Yuxin Wu and Kaiming He.
\newblock Group normalization.
\newblock In {\em European Conference on Computer Vision}, 2018.

\bibitem{wu2018unsupervised}
Zhirong Wu, Yuanjun Xiong, Stella~X Yu, and Dahua Lin.
\newblock Unsupervised feature learning via non-parametric instance
  discrimination.
\newblock In {\em IEEE Conference on Computer Vision and Pattern Recognition},
  2018.

\bibitem{xiao2018simple}
Bin Xiao, Haiping Wu, and Yichen Wei.
\newblock Simple baselines for human pose estimation and tracking.
\newblock In {\em European Conference on Computer Vision}, 2018.

\bibitem{crf}
Bo Yang and Ram Nevatia.
\newblock An online learned {CRF} model for multi-target tracking.
\newblock In {\em IEEE Conference on Computer Vision and Pattern Recognition},
  2012.

\bibitem{trackrcnn}
Linjie Yang, Yuchen Fan, and Ning Xu.
\newblock Video instance segmentation.
\newblock In {\em IEEE International Conference on Computer Vision}, 2019.

\bibitem{bdd100k}
Fisher Yu, Haofeng Chen, Xin Wang, Wenqi Xian, Yingying Chen, Fangchen Liu,
  Vashisht Madhavan, and Trevor Darrell.
\newblock Bdd100k: A diverse driving dataset for heterogeneous multitask
  learning.
\newblock In {\em IEEE Conference on Computer Vision and Pattern Recognition},
  2020.

\bibitem{poi}
Fengwei Yu, Wenbo Li, Quanquan Li, Yu Liu, Xiaohua Shi, and Junjie Yan.
\newblock {POI:} multiple object tracking with high performance detection and
  appearance feature.
\newblock In {\em European Conference on Computer Vision Workshop}, 2016.

\bibitem{zhang2020fair}
Yifu Zhang, Chunyu Wang, Xinggang Wang, Wenjun Zeng, and Wenyu Liu.
\newblock Fairmot: On the fairness of detection and re-identification in
  multiple object tracking.
\newblock {\em arXiv preprint arXiv:2004.01888}, 2020.

\bibitem{centertrack}
Xingyi Zhou, Vladlen Koltun, and Philipp Krähenbühl.
\newblock Tracking objects as points.
\newblock In {\em European Conference on Computer Vision}, 2020.

\bibitem{tap}
Zongwei Zhou, Junliang Xing, Mengdan Zhang, and Weiming Hu.
\newblock Online multi-target tracking with tensor-based high-order graph
  matching.
\newblock In {\em International Conference on Pattern Recognition}, 2018.

\end{thebibliography}
}

\end{document}